\documentclass{article}

\usepackage{arxiv}

\usepackage[utf8]{inputenc}          
\usepackage[T1]{fontenc}             
\usepackage{hyperref}                
\usepackage{url}                     
\usepackage{booktabs}                
\usepackage{amsfonts, amsmath}       
\usepackage{nicefrac}                
\usepackage{microtype}               
\usepackage{lipsum}		             
\usepackage{graphicx}
\usepackage{subcaption}
\usepackage{natbib}
\usepackage{doi}

\title{Storm Surge Modeling in the AI Era: Using LSTM-based Machine Learning for Enhancing Forecasting Accuracy}

\author{ \href{https://orcid.org/0000-0002-0107-3127}{\includegraphics[scale=0.06]{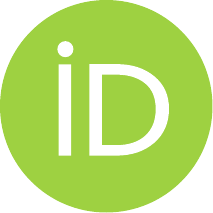}\hspace{1mm}Stefanos Giaremis} \\
	Department of Physics\\
	Aristotle University of Thessaloniki\\
	Thessaloniki, 54124, Greece \\
	\texttt{sgiaremi@physics.auth.gr} \\
	\And
	\href{https://orcid.org/0009-0000-4687-1416}{\includegraphics[scale=0.06]{orcid.pdf}\hspace{1mm}Noujoud Nader} \\
	Louisiana State University\\
	Baton Rouge, 70803, LA, USA \\
	\texttt{nnader@lsu.edu} \\
	\And
    \href{https://orcid.org/0000-0001-7273-0684}{\includegraphics[scale=0.06]{orcid.pdf}\hspace{1mm}Clint Dawson} \\
	Oden Institute for Computational \\
    Engineering and Sciences \\
	The University of Texas at Austin \\
    Austin, TX 78712, USA \\
	\texttt{clint@oden.utexas.edu} \\
	\And
    \href{https://orcid.org/0000-0002-8712-2806}{\includegraphics[scale=0.06]{orcid.pdf}\hspace{1mm}Hartmut Kaiser} \\
	Louisiana State University\\
	Baton Rouge, LA 70803, USA \\
	\texttt{hkaiser@cct.lsu.edu} \\
    \And
    Carola Kaiser \\
	Louisiana State University\\
	Baton Rouge, LA 70803, USA \\
	\texttt{ckaiser@cct.lsu.edu} \\
    \And
    \href{https://orcid.org/0009-0000-6220-9385}{\includegraphics[scale=0.06]{orcid.pdf}\hspace{1mm}Efstratios Nikidis} \\
	Department of Physics\\
	Aristotle University of Thessaloniki\\
	Thessaloniki, 54124, Greece \\
	\texttt{enikidis@physics.auth.gr} \\
}

\date{}

\renewcommand{\undertitle}{\textit{arXiv} preprint}
\renewcommand{\shorttitle}{LSTM-based Machine Learning for Bias Correction in Storm Surge Modeling}

\hypersetup{
pdftitle={LSTM-based Machine Learning for Bias Correction in Storm Surge Modeling},
pdfsubject={q-bio.NC, q-bio.QM},
pdfauthor={Stefanos Giaremis, Noujoud Nader, Hartmut Kaiser, Carola Kaiser, Efstratios Nikidis},
pdfkeywords={Long-Short Term Memory (LSTM) Networks, Machine Learning, Storm Surge Modeling, Bias Correction},
}

\begin{document}
\maketitle

\begin{abstract}
	Physics simulation results of natural processes usually do not fully capture the real world. This is caused for instance by limits in what physical processes are simulated and to what accuracy. In this work we propose and analyze the use of an LSTM-based deep learning network machine learning (ML) architecture for capturing and predicting the behavior of the systemic error for storm surge forecast models with respect to real-world water height observations from gauge stations during hurricane events. The overall goal of this work is to predict the systemic error of the physics model and use it to improve the accuracy of the simulation results \textit{post factum} (i.e., to correct the model bias).
    We trained our proposed ML model on a dataset of 61 historical storms in the coastal regions of the U.S. and we tested its performance in bias correcting modeled water level data predictions from hurricane Ian (2022). We show that our model can consistently improve the forecasting accuracy for hurricane Ian --- unknown to the ML model --- at all gauge station coordinates used for the initial data. Moreover, by examining the impact of using different subsets of the initial training dataset, containing a number of relatively similar or different hurricanes in terms of hurricane track, we found that we can obtain similar quality of bias correction by only using a subset of six hurricanes. This is an important result that implies the possibility to apply a pre-trained ML model to real-time hurricane forecasting results with the goal of bias correcting and improving the produced simulation accuracy.
    The presented work is an important first step in creating a bias correction system for real-time storm surge forecasting applicable to the full simulation area. It also presents a highly transferable and operationally applicable methodology for improving the accuracy in a wide range of physics simulation scenarios beyond storm surge forecasting.  
\end{abstract}


\section{Introduction}\label{sec:introduction}
Tropical cyclones, among other severe storms, often cause major flooding events and constitute the most frequent weather and climate related disasters in the U.S., causing the highest amount of financial costs and human losses (\cite{NCEI2023}). Climate change-induced sea level rise can also amplify the severity of such disasters (\cite{Yin2020}). State of the art high fidelity physics-based models, such as the ADvanced CIRCulation (ADCIRC) model, coupled with the Simulating WAves Nearshore model (SWAN) (\cite{Westerink1992, luettich1992, Dietrich2011}) are commonly used for both operational forecasting and analyzing, in retrospect, the behavior of severe hurricane events. Despite their strengths, such  models often require immense computational resources to achieve the desired level of accuracy. Moreover, since the evolution and impact of storms can differ widely depending on the geographic landscape, track direction and wind intensity, limitations and biases are often introduced in such deterministic forecasts. 
To overcome these issues, ensemble forecasting is commonly used in operational situations (\cite{Hllt2015, Ayyad2022, RezuanulIslam2023}); however, the parallel execution of multiple scenarios requires even more computational power. On the other hand, probabilistic approaches are explored as a more computationally efficient means of mitigating uncertainties propagation and rapid risk assessment (\cite{Resio2017, Jung2023}).

In this context, Machine Learning (ML) techniques are another increasingly popular approach for improving the flexibility, and thus potentially enhancing the practical applicability of real-time forecasting frameworks. One such approach involves the so-called surrogate modeling, which is the development of parametric models, fitted to data produced from highly accurate physics-based models or meteorological observations. Due to the ability of ML-based methods to capture and reproduce underlying non-linear relationships, their use for this task has become extremely popular in the last few years (\cite{Qin2023}). Recently in this direction, \cite{Pachev2023} demonstrated the promising performance of Artificial Neural Networks (ANN) and gradient tree boosting in predicting peaks in storm surge, as they found the accuracy of their framework to be comparable to ADCIRC but orders of magnitude faster. Their model was trained on several storm characteristics, accounting also for the complex bathymetry in certain regions, produced by either synthetic or historical storms simulations with ADCIRC and tested in against data from Hurricane Harvey (2017) and Hurricane Ike (2008) in Texas and Typhoon Merbok (2022) in Alaska. \cite{Mafi2017} used Support Vector Machine (SVM), Random Forests (RF) and ANN models, trained on buoy stations data from hurricanes Ike (2008) and Gustav (2008) to predict significant wave height. While the authors report improved predictions compared to previous studies due to using more input data, and their model yielded accurate predictions on non-hurricane events, it was found to underestimate significant wave height in the hurricane-affected part of the time period. \cite{Lee2021} proposed an ML architecture based on Convolutional Neural Networks (CNN), Principal Component Analysis (PCA) and k-means clustering. They trained their model on ADCIRC-based synthetic storms data on Chesapeake Bay and tested their prediction against historical data from hurricanes Isabel (2003),  Irene (2011) and Sandy (2012) on the same region. In a previous comparative analysis, \cite{AlKajbaf2020} also demonstrated that ANN along with Gaussian Process Regression (GPR) models performed with consistent accuracy in predicting storm surge height, despite the magnitude of the predicted value. A similar approach was also used in the context of a surrogate model for Computer Fluid Dynamics (CFD) simulations, for predicting storm surge and wave forces on elevated coastal buildings, by using K-Nearest neighbor (KNN), Stochastic gradient descent (SGD) and SVR models (\cite{Moeini2023}). 

Since such surrogate models essentially involve the prediction of time-dependent quantities, the use of Recurrent Neural Networks (RNN), and in particular Long-Short Term Memory (LSTM) based ML architectures are also recently explored in this context, due to their broad success in predicting patterns in sequential data (\cite{yu2019review, graves2013generating}). \cite{Ahn2022} used LSTM models trained on  31-years global hindcasts produced by WAVEWATCH III (WWIII) (\cite{WWIII, WWIII-hindcast}) for regional wave height forecasts in North Pacific, Gulf of Mexico, and North Atlantic. They reported reasonable predicting accuracy for nowcasting and short-period forecasting, with predicting performance limitations and systemic biases occurring when forecast time was increased up to 24h. \cite{Luo2022} trained LSTM-based models with hourly 1-year buoy stations data from hurricane-prone areas in the Atlantic ocean and found that bidirectional LSTM (Bi-LSTM) architectures with attention layers outperformed other the LSTM-based candidates. The use of Bi-LSTM was also shown by \cite{Bai2021} to lead to reasonable sea level predictions in data from five tidal stations in normal weather conditions. The performance of an encoder-decoder LSTM model, trained on meteorological data observations from 6 hurricanes in Fort Myers-FL for predicting storm surge in a single gauge station in lead times from 1 to 12h was assessed by \cite{Bai2022}. On a global scale analysis, \cite{Tiggeloven2021} trained four different ML models in meteorological data from the European Center for Medium-Range Weather Forecast’s (ECMWF) ERA5 global atmospheric reanalysis (\cite{Hersbach2020}) for predicting storm surge levels in 738 gauge stations from the Global Extreme Sea Level Analysis Version 2 (GESLA-2) dataset (\cite{GESLA}), and compared them with the respective observed values. They found that their LSTM model outperformed ANN, CNN and Convolutional LSTM (ConvLSTM) architectures. Moreover, the use of LSTM networks for predicting wind waves from buoy stations data was explored by \cite{Wei2021} with promising results.

Another approach for integrating ML techniques in forecast systems is as a post-processing bias correction component. Instead of trying to reproduce the behaviour of complex multi-component physical systems, this approach harvests the power of ML in predicting non-linear dependencies for the post-simulation mitigation of the systemic error of physics-based numerical models. So far, traditional bias correction approaches were focused on statistical methods such as linear/variance scaling and quantile mapping (\cite{Li2019}). However, \cite{Wang2022} demonstrated that the computer-vision inspired, Super Resolution Deep Residual Network (SRDRN) ML architecture (\cite{Wang2021}) outperformed other stochastic methods such as the quantile delta mapping approach, for bias corrections in temperature predictions from 20 coupled general circulation models. Their experiments were performed for the southeast U.S. and the Gulf of Mexico, and they used the ECMWF ERA5 dataset as reference for the observed data. \cite{Sun2022} used the BU-net architecture, which is based on the U-Net deep model by \cite{Ronneberger2015} (also initially designed for computer vision applications) with additional batch normalization layers (\cite{Ioffe2015}), for bias correcting significant wave height predictions from WWIII. They used 4-years data from the Northwest Pacific Ocean. \cite{Ellenson2020} also implemented a ML bias correction model to WWIII significant wave height predictions by using bagged regression trees in 24 h forecasts in the California-Oregon border.  The use of Multilayer Perceptrons (MLP) for improving the accuracy of significant wave height and 10-m wind intensity predictions by the Global ocean Wave Ensemble forecast System (GWES) from the U.S. National Centers for Environmental  Prediction (NCEP) (\cite{Alves2013}), with altimeter data as reference, was explored by \cite{Campos2020}. RNNs based on the LSTM architecture were also explored for bias correcting numerical predictions for sea surface temperature (\cite{Fei2022}), but also for the Madden-Julian Oscillation amplitude (\cite{Kim2021}).

In this work, we assess the performance of an LSTM-based ML architecture as a post-processing, bias correction model, focused on improving the accuracy of storm surge prediction in hurricane events. Our model was trained on a dataset of 61 historical storms that impacted the east coast of the US, including numerical water level predictions by ADCIRC, produced in the CERA framework by \cite{CERA2023}. Our approach is applicable to real-time scenarios, since the bias correction model is only trained once, so the pre-trained model can be applied on top of the physics-based model without significant computational cost, and also transferable to forecasting of other quantities (such as wind) or in other regions, since it is not explicitly dependent on any specific parameters of the numerical model or the geographical region we considered in our experiment. The data and methods we employed, including a description of the ML architecture and the training scenarios we considered, are described in Section \ref{sec:data_and_methodology}. The results of our proposed ML bias correction model are presented in Section \ref{sec:results_and_discussion}. Finally, the summary and the concluding remarks of this work are included in Section \ref{sec:conclusions}.

\section{Data and Methodology}\label{sec:data_and_methodology}
\subsection{Overview}

The procedure followed in this work involved the systematic extraction of the offsets between the modeled and observed water height time series from each gauge station in the available data, and the subsequent data processing and reshaping, the selection and training of the ML model and the hypertuning of its parameters, and finally the evaluation of its performance. The offsets time series for each gauge station is defined as: 

\begin{align}
    H_\text{offsets}(t) = H_\text{modeled (physics-based model)}(t) - H_\text{observed (gauge stations)}(t) 
\end{align} \label{eq:offsets}
where $H_\text{offsets}(t)$, $H_\text{modeled (physics-based model)}(t)$ and $H_\text{observed (gauge stations)}(t)$ are the offsets and the modeled and observed water levels in ft., respectively, at time $t$.

\begin{figure}[h!]
	\centering
		\includegraphics[width=1\textwidth]{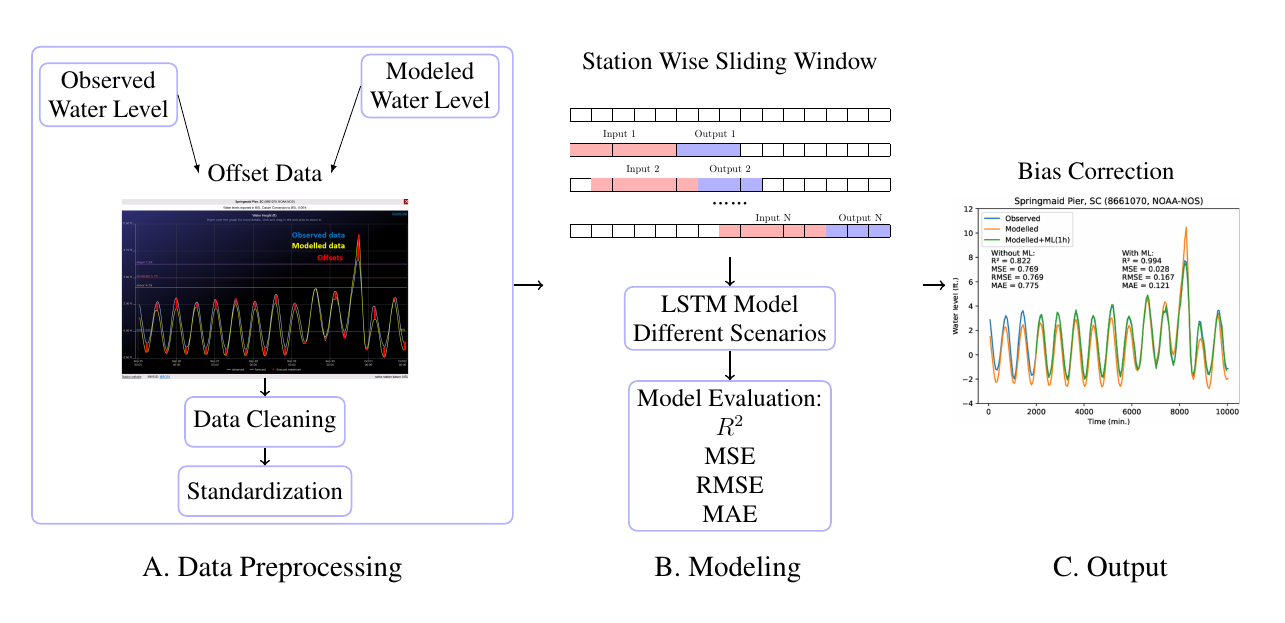}
	  \caption{Overview of the methodology framework: (A) Data Pre-processing phase includes offset extraction using Eq. \ref{eq:offsets}, data cleaning, and standardization. (B) Modeling phase includes training/test data preparation with the sliding window approach for the offsets time series at each station. Different scenarios were analyzed and evaluated with several commonly used regression evaluation metrics. A single ML model is used in each scenario for all the considered stations.} (C) Output phase includes the final use of the pre-trained model for bias correction.
   \label{fig:pipeline}
\end{figure}

This process is schematically represented in Figure \ref{fig:pipeline}. It is composed of three parts, including data pre-processing,
modeling and output. The first part in Figure \ref{fig:pipeline}A includes offset extraction using Eq. \ref{eq:offsets} data cleaning, and standardization. The second part in Figure \ref{fig:pipeline}B consists of preparing the training/test data using the sliding window approach for the time series offsets samples at each station. Different scenarios were analyzed and evaluated using commonly employed regression evaluation metrics. Subsequently, the model can
be finally applied for bias correction, as shown in Figure \ref{fig:pipeline}C. A detailed description of each step, along with the choice of training and testing data in different scenarios, is included in the following sections.

\subsection{Data}
\label{ssec:data}

Water level data from 61 historical storms were obtained from the Historical Storm Surge Archive by \cite{CERA2023} and visualized via the built-in interface with the \cite{CERA} website. Data for each hurricane include physics-based modeled values, produced by ADCIRC (\cite{Westerink1992}), and observations from National Oceanic and Atmospheric Administration (\cite{NOAA}) and coastal United States Geological Survey (\cite{USGS}) gauge stations, at an hourly resolution. Offsets, as defined by Eq. \ref{eq:offsets}, were extracted by calculating the difference between observed and modeled water level at each hourly time interval, leading to the construction of an offset time series for each available station in each hurricane of the dataset. The full list of the 61 hurricanes used in this work is reported in Supplementary Information, Table 1.

\begin{figure}[h!]
     \centering
     \begin{subfigure}[b]{\textwidth}
         \centering
         \includegraphics[scale=0.5]{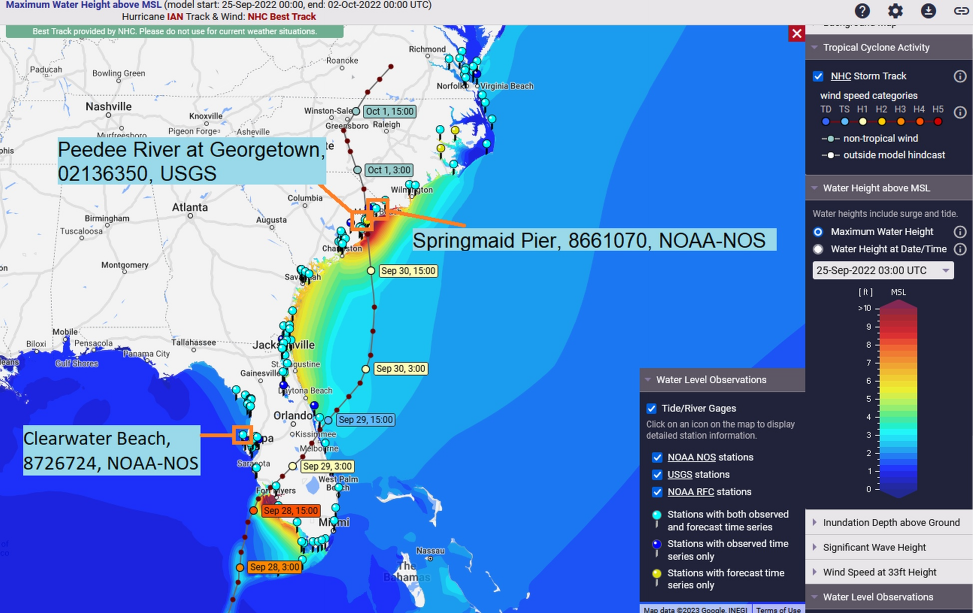}
         \caption{}
         \label{fig:methods_cera_1}
     \end{subfigure}
     \hfill
     \begin{subfigure}[b]{\textwidth}
         \centering
         \includegraphics[scale=0.5]{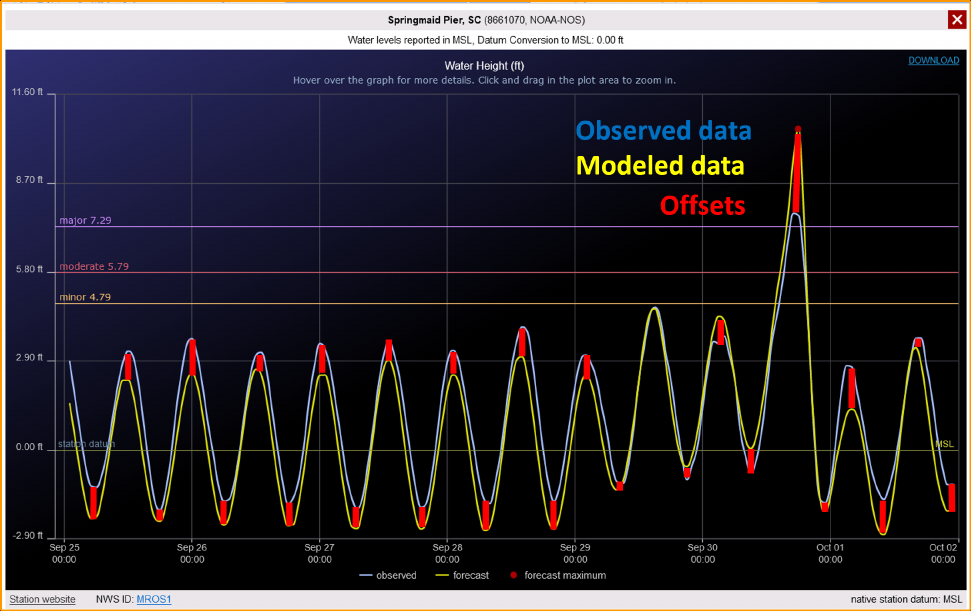}
         \caption{}
         \label{fig:methods_cera_2}
     \end{subfigure}
     \caption{(a) Visualization of hurricane Ian water levels in contour, along with the National Hurricane Center best track, available gauge stations data and the selection of three stations used for the demonstration of the performance of the ML model, within the CERA framework (\cite{CERA}, \cite{CERA2023}). (b) Visualization of the observed and modeled water level, along with the corresponding offset time series (as defined by eq. \ref{eq:offsets}, from an example gauge station (Springmaid Pier) during hurricane Ian, as obtained from the interactive CERA framework (\cite{CERA}, \cite{CERA2023}).}.
     \label{fig:methods_cera}
\end{figure}

To examine the impact of using training data from hurricanes with different characteristics to the performance of the ML bias correction model, six different scenarios were considered (Table \ref{tbl:scenarios}). Hurricane Ian (2022) was set as the main point of interest in this work and was used as target storm for comparison between the different scenarios. Thus hurricane Ian was used as the test set in all cases. The first scenario ("Ian") involved using offsets data only from hurricane Ian (2022), with the chronologically earliest 75\% of data used to train the ML model and the latest 25\% as a test set for its evaluation. The second scenario ("All") involved using offsets data from 60 out of the 61 hurricanes as training set, and data from hurricane Ian (2022) as test set. The third and fourth scenarios involved considering as training sets offsets data from either one similar ("1 similar") or one different ("1 different") hurricane, compared to Ian, in terms of hurricane tracks. Hurricane Charley (2004) was used in the third ("1 similar") and hurricane Harvey (2017) was used in the fourth scenario ("1 different"), respectively. Similarly, six hurricanes with relatively similar tracks compared to Ian (Charley (2004), Wilma (2005),  Hermine (2016), Irma (2017), Eta (2020) and Elsa (2021)), and six hurricanes with different tracks compared to Ian (Sandy (2012), Matthew (2016), Harvey (2017), Michael (2018), Florence (2018) and Ida (2021)) were considered for the training sets in the fifth and sixth scenarios ("6 similar" and "6 different"), respectively. In all cases, no information from the test data (i.e., 25\% of the chronologically latest offset data from hurricane Ian in the first scenario, and all offset data from hurricane Ian in the rest) was used in the training process of the models. The amount of available hourly offsets data for each scenario are shown in Table \ref{tbl:scenarios}. A screenshot from the visualization of hurricane Ian water levels and the National Hurricane Center best track via the CERA website (\cite{CERA}), along with a selection of three stations used later for the demonstration of the performance of the ML model, are shown in Figure \ref{fig:methods_cera_1}. A screenshot from the interactive visualization of observed and modeled water levels time series in an example gauge station (Springmaid Pier) from the same source, along with a graphical representation of the respective offsets time series (eq. \ref{eq:offsets}) is shown in Figure \ref{fig:methods_cera_2}.

\begin{table}[h]
\caption{Hurricanes and the total amount of hourly offsets data used for training in each of the considered scenarios.}\label{tbl:scenarios}
\begin{tabular}{lp{7cm}c}
\toprule
Scenario  & Hurricanes & Amount of hourly offsets training data \\ 
\midrule
1: Ian & Ian (2022) & 26111 \\
2: All & Full dataset (see S.I., Table I)  & 931436 \\
3: 1 similar & Charley (2004) & 14581\\
4: 1 different & Harvey (2017) & 18334\\
5: 6 similar & Charley (2004), Wilma (2005), Hermine (2016), Irma (2017), Eta (2020), Elsa (2021) & 75461\\
6: 6 different & Sandy (2012), Matthew (2016), Harvey (2017), Michael (2018), Florence (2018), Ida (2021) & 126160
\\

\bottomrule
\end{tabular}
\end{table}

\subsection{Machine Learning Model} 
In this section, we introduce our ML model for predicting storm surge forecast errors. First, we provide an overview of the LSTM networks, the cornerstone of our ML approach. Following that, we show in details our proposed LSTM network architecture, including its structure and configuration. Finally, we describe the data pre-processing steps used to transform the modeled and observed water levels data from NOAA and USGS gauge stations into a format suitable for input into our predictive model. 

\subsubsection{Long Short-Term Memory}
One-step-ahead forecasting in time series necessitates access to not only the most recent data but also past data. The RNN model, with its inherent self-feedback mechanism within the hidden layer, offers an advantage in addressing long-term dependence issues, although practical implementation poses challenges (\cite{bengio1994learning}). To tackle the gradient vanishing problem encountered in RNNs, Sepp Hochreiter and Jurgen Schmidhuber introduced the LSTM model in 1997 (\cite{hochreiter1997long}), a model further refined and popularized by Alex Graves (\cite{graves2013generating}). The LSTM unit is composed of a memory cell that stores and updates information through the operation of three distinct gates: the input gate, the forget gate, and the output gate. \\

At time step t, an input denoted as $x_t$ is provided, and the LSTM block receives the hidden state from the preceding time step $h_{t-1}$. Subsequently, the hidden state, represented as $h_t$, is computed using the following process:\\
\begin{enumerate}
    \item The first step in the LSTM process is to decide which data should be removed from the cell state using the forget gate $f_t$. This data selection process is orchestrated by the sigmoid function, which operates on both $h_{t-1}$ and the current input $x_t$. Furthermore, the sigmoid function determines which components from the previous output should be omitted. The result is then a vector with values ranging from 0 to 1, each corresponding to elements within the cell state (\(C_{t-1}\)).
\begin{equation}
 f_t = \sigma(W_f \cdot [h_{t-1}, x_t] + b_f).
\end{equation}
Herein, $\sigma$ is the sigmoid activation function. $W_f$ and $ b_f $ are the weight matrix and bias for the forget gate, respectively.
\\
\item The following phase involves the selection and retention of information from the new input $(x_t)$ within the cell state, as well as updating the cell state itself. This process comprises two essential components: first, the sigmoid layer determines whether the new information should be preserved or disregarded (resulting in a binary decision of 0 or 1); and second, the hyperbolic tangent ($tanh$) function assigns weights to the data that passes through, signifying their relative significance within a range of -1 to 1. These two computed values are then multiplied together to effectuate the update of the new cell state. This new updated memory is subsequently combined with the previous memory, $C_{t-1}$, resulting in the updated cell state, $C_t$.

\begin{align}
i_t &=\sigma(W_i \cdot [h_{t-1}, x_t] + b_i)\\
\widetilde{C}_t &= \text{tanh}(W_c \cdot [h_{t-1}, x_t] + b_c)\\
C_t &= f_t * C_{t-1} + i_t * \widetilde{C}_t.
\end{align}
Where, "$*$" represents dot product, the cell states at times \(t - 1\) and \(t\) are denoted by \(C_{t-1}\) and \(C_t\), respectively, while the weight matrices and bias are represented by \(W\) and \(b\).
\\
\item Finally, the output values ($h_t$) in the last step are based on the output cell state ($o_t$), however they are a filtered version. First, a sigmoid layer select which parts of the cell state  are sent to the output. Subsequently, the output of the sigmoid gate (\(o_t\)) is multiplied by the new values generated by the $tanh$ layer from the cell state (\(C_t\)), values ranging between -1 and 1.
\begin{align}
o_t &= \sigma(W_o \cdot [h_{t-1}, x_t] + b_o)\\
h_t &= o_t*\text{tanh}(C_t).
\end{align}
Here, the output gate's weight matrices and bias are denoted by $W_o$ and $b_o$, respectively.
\end{enumerate}

\subsubsection{Convolution Layer}
To consider spatial dependence in our time series offsets data, we have used a convolutional layer \cite{yang2015deep}. This process is used in convolution neural networks (CNN). CNN with a single layer extracts features from the input signal through a convolution operation of the signal with a filter (or kernel). The activation of a unit represents the result of the convolution of the kernel with the input signal. By computing the activation of a unit on different regions of the same input (using a convolutional operation), it is possible to detect patterns captured by the kernels, regardless of where the pattern occurs. 

The application of the convolution operator depends on the input dimensionality. With a one-dimensional temporal sequence often a 1D kernel is used in a temporal convolution (\cite{kiranyaz20211d}). In the 1D domain, a kernel can be viewed as a filter, capable of removing outliers, filtering the data or acting as a feature detector, defined to respond maximally to specific temporal sequences within the timespan of the kernel. Formally, extracting a feature map using a one-dimensional convolution operation is given by:
\begin{equation}
A_j = \sigma\left(\sum_{i=1}^{N} I_i \ast K_{i,j} + B_j\right)
\label{eq:convolution_layer}
\end{equation}
Herein, the input time series signal is denoted by $I_i$, $K_{i,j}$ is the kernel matrix , $B_j$ is the bias, $\sigma$ is the non-linear activation and  output matrix $A_j$.
The non-linear activation function used for this layer was the ReLU (see Section \ref{sssec:activation_func}).

\subsubsection{Fully Connected Layer}
The fully connected layer, also known as the dense layer, constitutes a pivotal element within neural network architectures, facilitating comprehensive interconnections between input and output data.
Fully connected layers in neural networks are responsible for managing flattened feature vectors and ultimately producing final predictions (\cite{lecun2015deep}). A fully connected layer operation can be represented as: 
\begin{equation}
y = \sigma(W \cdot x + b)
\label{eq:fully_connected_layer}
\end{equation}
where the activation function is $\sigma$, the weight matrix is $W$, the bias vector is $b$, the output is $y$, and the input vector is $x$.

\subsubsection{Activation Functions}
\label{sssec:activation_func}

\begin{itemize}
    \item \textbf{The rectified linear unit ReLU:}\\
 The rectified linear unit (ReLU) is widely used as a non-linear activation functions in neural network. ReLU is defined as:  $\sigma_{\textbf{ReLU}}(x)=max(0,x)$.
    \\
    \item \textbf{Linear Activation function:}\\
The linear activation function maintains a direct proportionality with its input. Unlike the binary step function, which lacks a gradient due to its absence of a component for $x$, the linear function addresses this limitation. This function, defined as 
$\sigma_{\textbf{linear}}=ax$, allows for the inclusion of $x$ in its formulation, enabling a continuous gradient for better learning and optimization in neural networks.
    \\
    \item \textbf{Sigmoid:}\\
As the most used activation function due to its non-linear behavior, the sigmoid function effectively transforms values within the range of 0 to 1. Its formulation \[\sigma_{\textbf{sigmoid}} = \frac{1}{1 + e^{-x}}\] ensures the conversion of input data into a bounded output, making it valuable in various neural network architectures for introducing non-linearity.
    \\
    \item \textbf{Tanh:}\\
 The hyperbolic tangent, or tanh, activation function is a popular choice in neural networks owing to its characteristics. Similar to the sigmoid function, tanh is non-linear, but it maps input values to a range between -1 and 1. This zero-centered nature allows tanh to produce outputs that are easier to work with in optimization processes, particularly in scenarios where data normalization is crucial \cite{sharma2017activation}. The tanh function is defined as 
 \begin{equation}
\sigma_{\textbf{tanh}} = \frac{e^{x} - e^{-x}}{e^{x} + e^{-x}}
 \end{equation}
\end{itemize}

\subsubsection{Structure}
\begin{figure}[h!]
	\centering
		\includegraphics[width=1\textwidth]{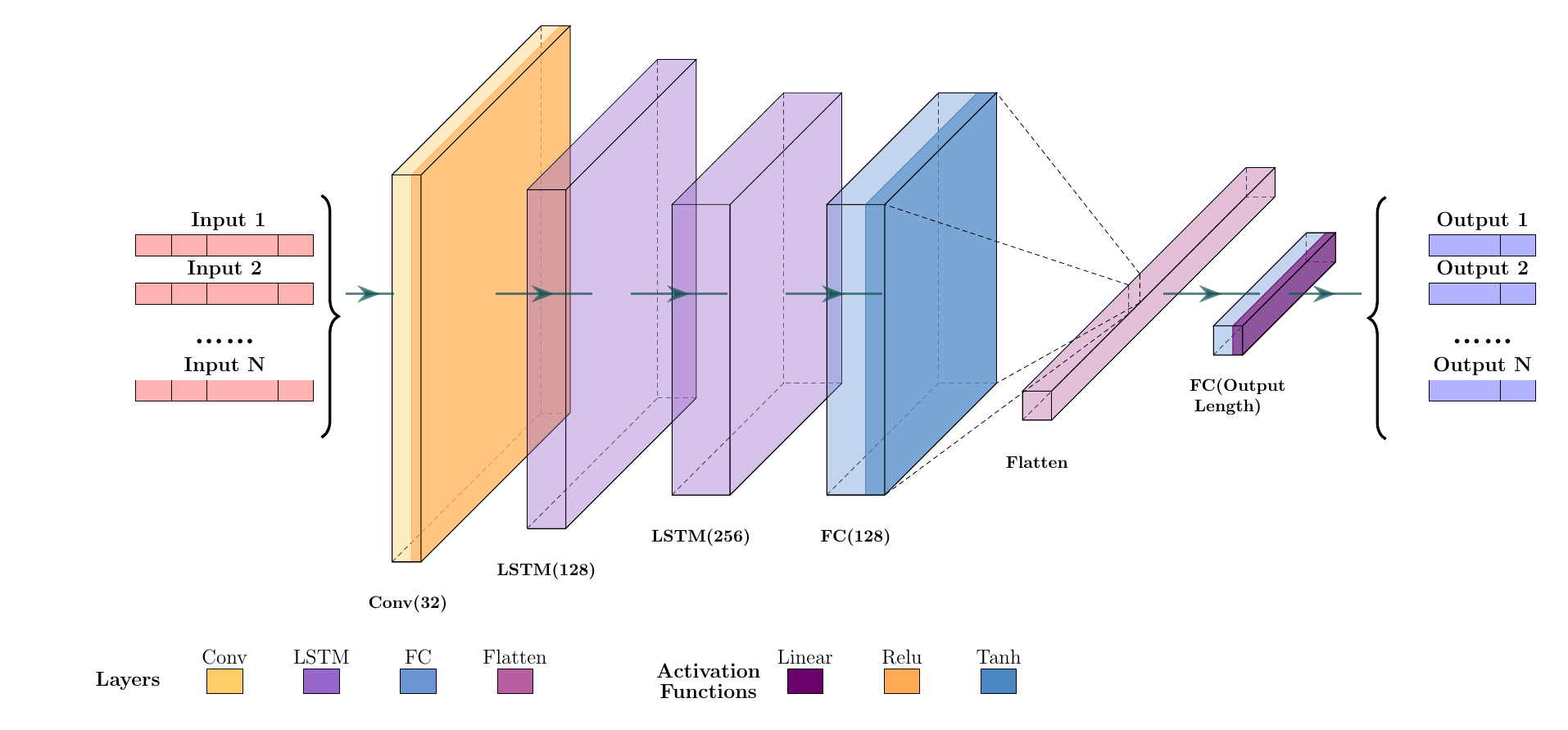}
	  \caption{Architecture of the proposed ML-model for offset prediction. The input samples are the offset time series windows. The architecture consist of a Convolution layer followed by multiple LSTM layers. The output of the model is the predicted offset window. }
   \label{fig:LSTM_network}
\end{figure}

In this study, we have developed a neural network primarily centered around Long Short-Term Memory (LSTM) networks. The input samples are the offset time series windows and the output of the model is the predicted offset window. More information on these terms, regarding data preparation via the sliding window approach, is included in Section \ref{sec:Windowing} However, as we are also interested in recognizing the significance of spatial dependencies in our dataset, thus we have integrated a convolutional layer into our architecture. As shown in Figure. \ref{fig:LSTM_network}, the first layer of our network is a convolutional layer with a  ReLU activation function for feature extraction. This layer uses 32 filters with 3 kernels. This is followed by two LSTM layers, each with 128 and 256 units, respectively, to effectively model long-term dependencies in the data. These layers play a crucial role in understanding and capturing complex temporal relationships.

To further enhance the capabilities of the model, we included a dense (fully connected) layer with 128 units and a hyperbolic tangent ($tanh$) activation function, which adds non-linearity to the network. The Flatten layer is employed to transform the multidimensional output into a flat vector, preparing it for the final prediction step. Finally, we use another dense layer with a linear activation function to produce our output, which represents the predicted value. The number of neurons in the last layer depends on the predicted (output) window length.

\subsubsection{Computational resources and model parameters}
The ML model described in the previous section was trained by using the Adam optimizer (\cite{Kingma2014}), 200 epochs and a batch size of 32. The learning set was set to the commonly used value of 0.001. The number of layers and neurons per each layer in the aforementioned model architecture was determined by employing both Bayesian optimization (\cite{Frazier2018}) and a brute-force grid search for selected values of hyperparameters, as implemented in \cite{pedregosa2011scikit} via the \texttt{BayesSearchCV} and \texttt{GridSearchCV} classes, respectively. The models were trained on NVidia® A-100® GPUs, except in scenario 1 (see Table \ref{tbl:scenarios}), in which Intel® Xeon® Platinum 8260 CPUs were used due to the small size of the dataset.

\subsubsection{Windowing and Data Preprocessing}
\label{sec:Windowing}
The offsets time series, discussed in Section \ref{ssec:data} require pre-processing before they can be utilized for training and evaluating our neural network model. This section elaborates on the steps involved in this data transformation and preparation. The initial data preparation step involves tackling the challenge of disparate data scales. To address this issue, we have employed the normalization process, which consists of subtracting the minimum value from the offset data and dividing by their range (defined by the difference between the maximum and minimum values). As a result, the transformed data distribution are in the range $[0, 1]$, as recommended by \cite{geron2022hands}. For this purpose, we have used the \texttt{MinMax} transformer from Scikit-Learn (\cite{pedregosa2011scikit}). The data for each station were divided into training and test sets, and afterwards they underwent normalization. The training of the \texttt{MinMax} scaler (i.e., the estimation of the minimum and maximum values required for the normalization) was performed only on the training set used in each case, to avoid data leakage. Subsequently, the training set was used to train a single model using the offsets data from all the considered stations, while the test set is used to evaluate the performance of the model in future offset forecasting. It is important to note that the test set was entirely unknown to the model during its training phase.

\begin{figure}[h!]
	\centering
		\includegraphics[width=1\textwidth]{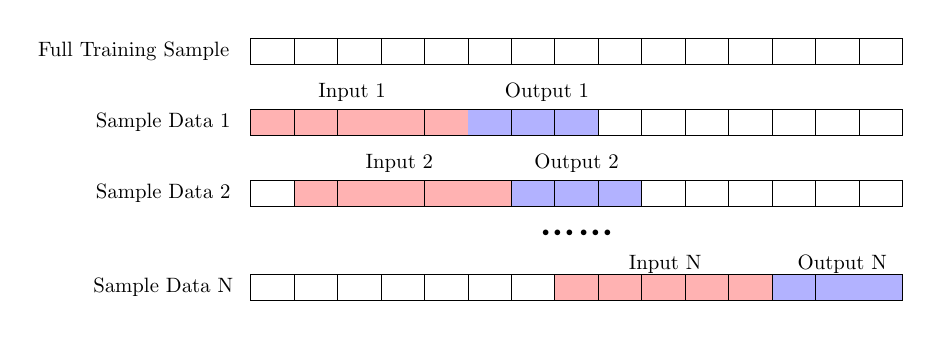}
	  \caption{The sliding window approach: Data samples processed from a time series training sample by creating input-output pairs.}
   \label{fig:sliding_window}
\end{figure}

After train/test splitting, both training and test sets data were subsequently split into windows using the windowing approach, which is a fundamental technique in time series forecasting using LSTM networks. It transforms continuous time series data into a supervised learning problem by creating input-output pairs from historical observations. This approach involves defining a fixed-size window that slides through the time series, creating input sequences with a set of consecutive past data points and their corresponding target value, typically the data point immediately following the window. LSTM, with its recurrent connections, is well-suited for this sequential data processing. Training the LSTM model with such input-output pairs allows it to capture and learn from temporal dependencies and patterns in the data, making it an effective tool for time series forecasting (\cite{Wei2021}). For instance, if our goal is to forecast offsets for the next \textit{Predicted Window} hours using data from the preceding \textit{Input window} hours, the input configuration for the LSTM model comprises the window of \textit{Input} hours, while the target corresponds to the window spanning \textit{Output} hours as shown in Figure \ref{fig:sliding_window}.
\subsubsection{Evaluation Metrics}
To assess the performance of the model, we employed four distinct metrics: Root Mean Square Error (RMSE), Mean Absolute Error (MAE), Mean Squared Error (MSE), and the R-squared coefficient (R²). These metrics are widely used in quantifying the accuracy and predictive capacity of such models.

Here are the definitions of these metrics:

- \textbf{Mean Squared Error (MSE)} computes the average of the squared differences between real values and predicted values. MSE is defined as:
\begin{equation}
\text{MSE} = \frac{1}{n} \sum_{i=1}^{n} (y_i - \hat{y}_i)^2
\end{equation}

- \textbf{Root Mean Square Error (RMSE)} measures the square root of the average of the squared differences between the real values ($y$) and the predicted values ($\hat{y}$). It is defined as:

\begin{equation}
\text{RMSE} = \sqrt{\frac{1}{n} \sum_{i=1}^{n} (y_i - \hat{y}_i)^2}.
\end{equation}

- \textbf{Mean Absolute Error (MAE)} calculates the average of the absolute differences between the real values and the predicted values. It is defined as:

\begin{equation}
\text{MAE} = \frac{1}{n} \sum_{i=1}^{n} \rvert y_i - \hat{y}_i \rvert
\end{equation}

- \textbf{R-squared coefficient (R²)} represents the proportion of the variance in the dependent variable that is predictable from the independent variables. It is a measure of goodness of fit and is given by:

\begin{equation}
\text{R²} = 1 - \frac{\sum_{i=1}^{n} (y_i - \hat{y}_i)^2}{\sum_{i=1}^{n} (y_i - \bar{y})^2}
\end{equation}
Herein,  $y$ represents the real offset values, $\hat{y}$ stands for the predicted offset values, $\bar{y}$ is the average of the observed values and $n$ denotes the total number of samples. These metrics collectively provide valuable insights into the model's performance and its ability to make accurate predictions.

\section{Results and Discussion}\label{sec:results_and_discussion}

\subsection{Scenario 1: ML bias correction using training data from a single hurricane}
\label{ssec:scenario1}

\begin{figure}[h!]
     \centering
     \begin{subfigure}[b]{0.45\textwidth}
         \centering
         \includegraphics[scale=1]{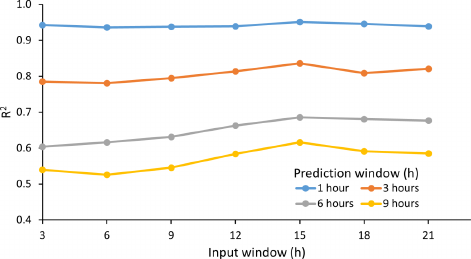}
         \caption{}
         \label{fig:R2_Ian_1}
     \end{subfigure}
     \begin{subfigure}[b]{0.45\textwidth}
         \centering
         \includegraphics[scale=1]{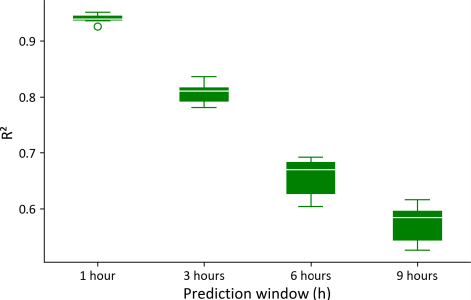}
         \caption{}
         \label{fig:R2_Ian_2}
     \end{subfigure}
     \begin{subfigure}[b]{0.45\textwidth}
         \centering
         \includegraphics[scale=1]{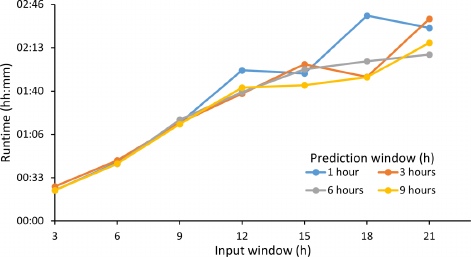}
         \caption{}
         \label{fig:R2_Ian_3}
     \end{subfigure}
     \caption{(a) Performance of the ML model in predicting offset data, as evaluated with the R$^2$ metric, as a function of input window size for different sizes of the prediction window. (b) Distribution of R$^2$ obtained in the considered range of input window sizes, as a function of the prediction window size. (c) Runtime required for the training of the model as a function of the input window, for different sizes of the prediction window. In all cases, the model was trained on the earliest 75\% of offsets data from hurricane Ian (2022), while the rest 25\% of the data of the same hurricane were used for testing its performance. 4 Intel® Xeon® Platinum 8260 processor cores were used for the training of the ML models. }
   \label{fig:R2_Ian}
\end{figure}

To demonstrate the applicability of the considered ML model in the current regression problem, we first examined its performance on a short scale dataset, involving data only from hurricane Ian. Data were sorted in chronological order and subsequently, the (chronologically) earliest 75\% of measurements were used to train the model and the rest were kept hidden from the model during training and were used only for testing its performance. The impact of varying the length of the input window to the performance of the model in the test set, for different prediction windows, is presented in Figure \ref{fig:R2_Ian}. The highest R$^2$ value for each prediction window, along with the corresponding input window, are presented in Table \ref{tbl:Results-Ian}. It is evident that optimal performance, reflected by a maximum value of R$^2$, is achieved when a 15h input window is used during the windowing pre-processing. Increasing input window also leads to a linear increase in the computational time required for the training of the model, as shown in Figure \ref{fig:R2_Ian_3}. The model is found to yield highly accurate results (R$^2$ > 0.9) when used to predict offsets 1h ahead, however its performance becomes limited with increasing prediction window length. The latter is reflected by an observed decrease in R$^2$ by $\sim 0.15$ for every 3 hours of increase in prediction window length. Moreover, from Figure \ref{fig:R2_Ian_2}, it is evident that the variance of R$^2$ increases with increasing prediction window length. This variance expresses the impact of optimizing the length of the input window, therefore these results demonstrate that in this scenario, the impact of optimizing the input window length is becoming increasingly pronounced the further ahead the model is required to predict. 

\begin{table}[h!]
\caption{Optimal performance (in R$^2$) and size of the input window required for achieving the optimal performance for the three considered sizes of the prediction window. The model was trained on the earliest 75\% of offsets data from hurricane Ian (2022), while the rest 25\% of the data of the same hurricane were used for testing its performance.}\label{tbl:Results-Ian}
\begin{tabular}{ccc}
Prediction window (h) & Optimal R$^2$ & Optimal input window (h)\\
\toprule
1 & 0.95 & 15 \\
3 & 0.84 & 15 \\
6 & 0.70 & 15 \\
9 & 0.62 & 15 \\
\bottomrule
\\
\end{tabular}
\end{table}

\subsection{Scenario 2: ML bias correction using the full hurricane dataset for training}
\label{ssec:scenario2}
The first approach, presented in Section \ref{ssec:scenario1}, demonstrated the feasibility of the proposed approach for bias correction, as a kind of a ``proof of concept" example. However, both the observed limited performance with increasing prediction window, and the fact that the applicability of such an approach in a real-time, real-world scenario is questionable (as the model would have to be trained in a a short amount of time to produce results in time, and subsequently re-trained in frequent intervals as new data would become available), in the second scenario, we followed the most conservative direction, by using all the available offsets data from past hurricanes from \cite{CERA2023} (see Table \ref{tbl:scenarios} and Supplementary Information, Table 1). Such an approach, as also the ones considered in the next scenarios, are highly applicable in real scenarios for applying post-simulation bias corrections ``on-the-fly", since the ML model intended to perform this task is only being trained once, and invoking the pre-trained model for applying the bias corrections requires an almost negligible computational cost (at least compared with the cost required for its training). Moreover, a significantly greater amount of data is used in this scenario. 

\begin{figure}[h!]
     \centering
     \begin{subfigure}[b]{0.325\textwidth}
         \centering
         \includegraphics[scale=0.37]{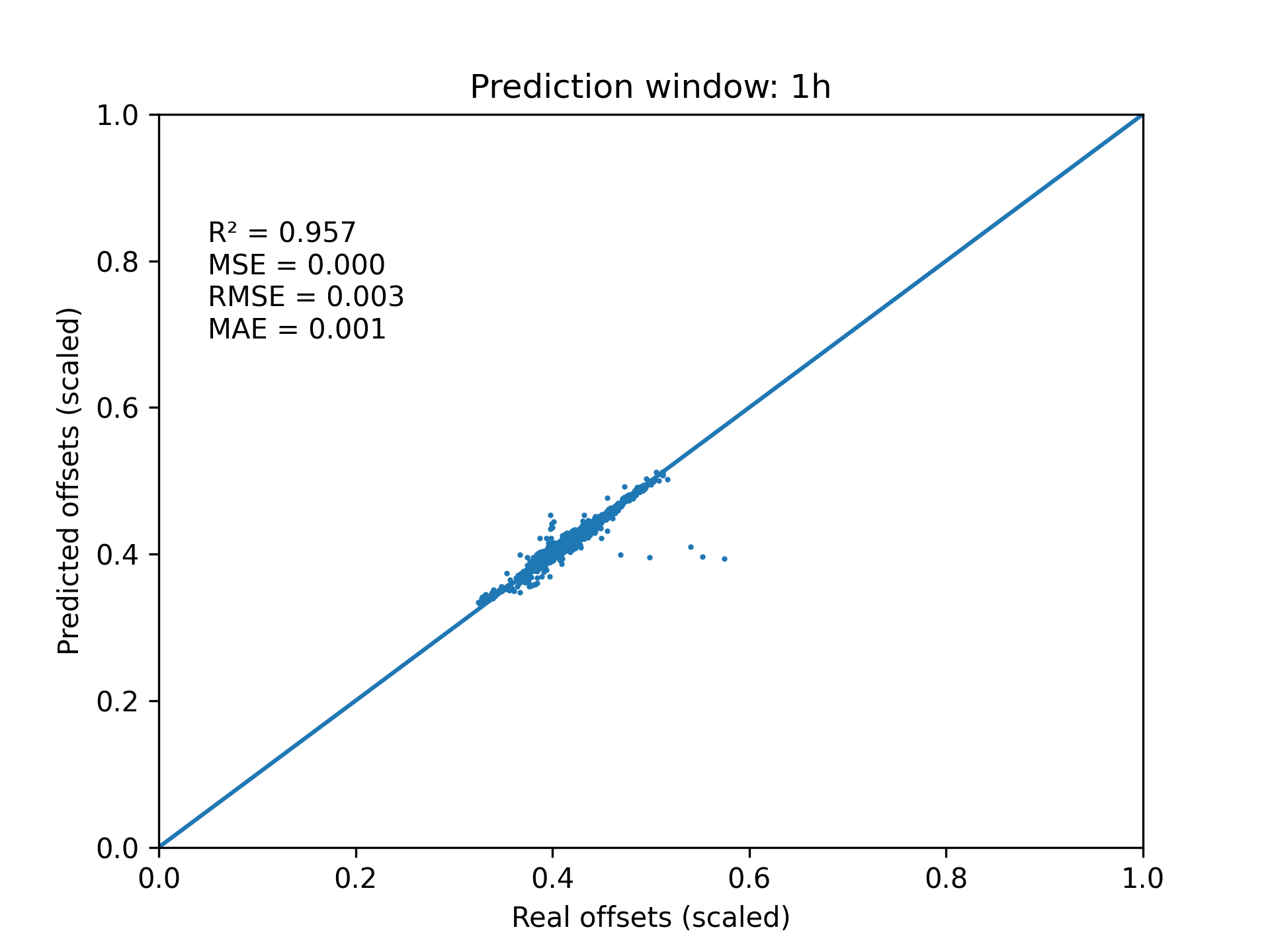}
         \caption{}
     \end{subfigure}
     \hfill
          \begin{subfigure}[b]{0.325\textwidth}
         \centering
         \includegraphics[scale=0.37]{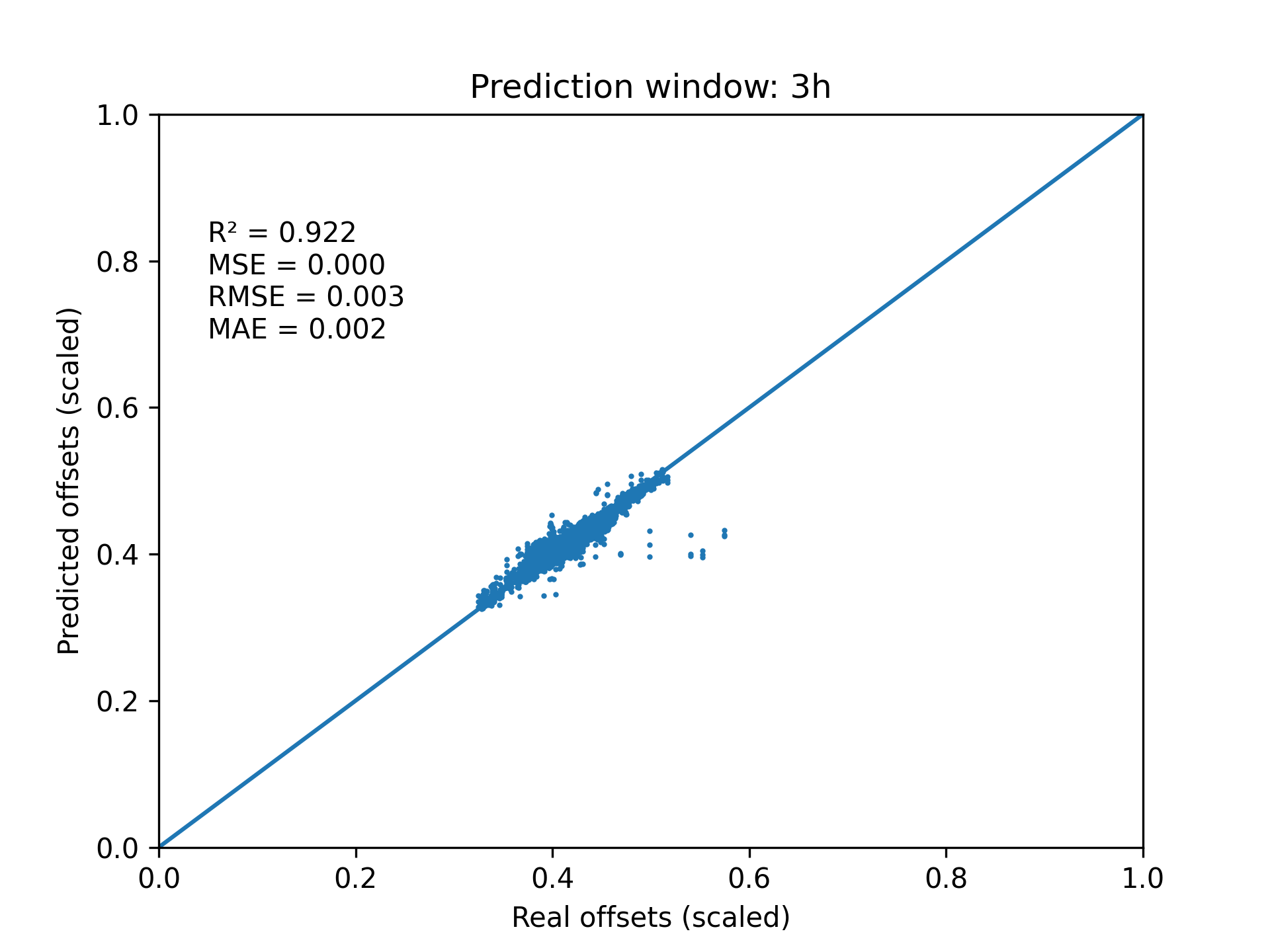}
         \caption{}
     \end{subfigure}
     \hfill
               \begin{subfigure}[b]{0.325\textwidth}
         \centering
         \includegraphics[scale=0.37]{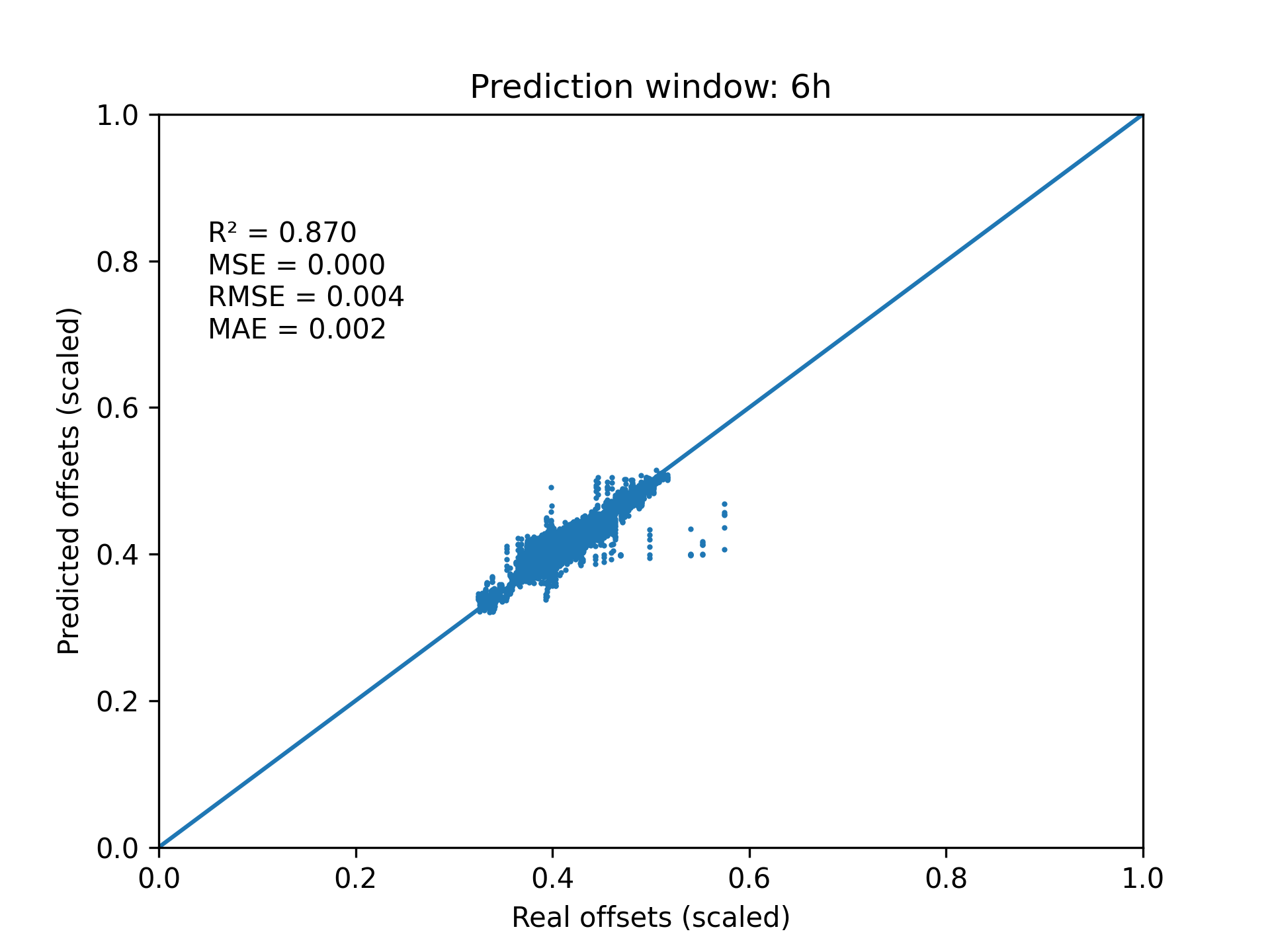}
         \caption{}
     \end{subfigure}
     \hfill
          \begin{subfigure}[b]{0.325\textwidth}
         \centering
         \includegraphics[scale=0.37]{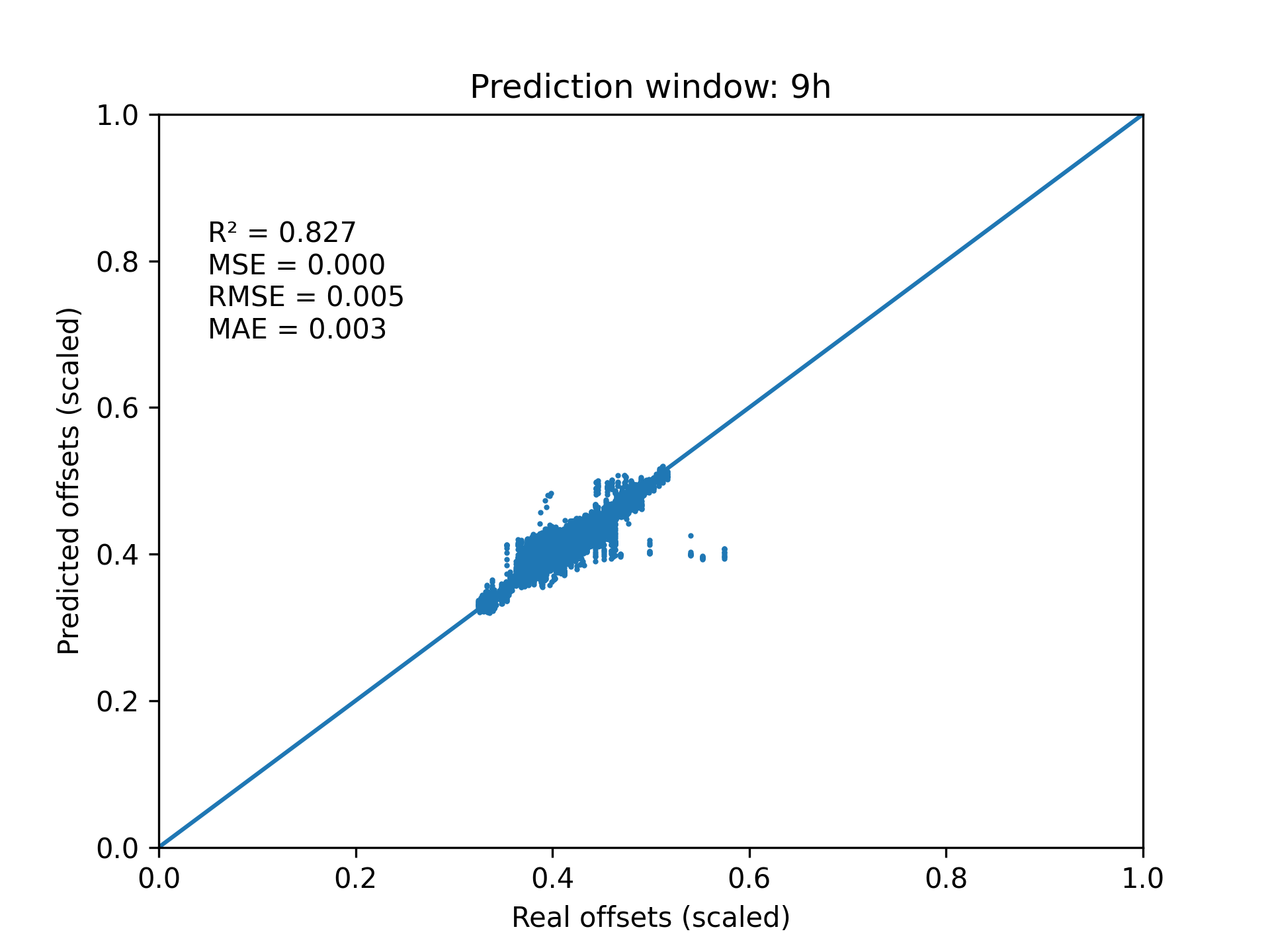}
         \caption{}
     \end{subfigure}
     \hfill
          \begin{subfigure}[b]{0.325\textwidth}
         \centering
         \includegraphics[scale=0.37]{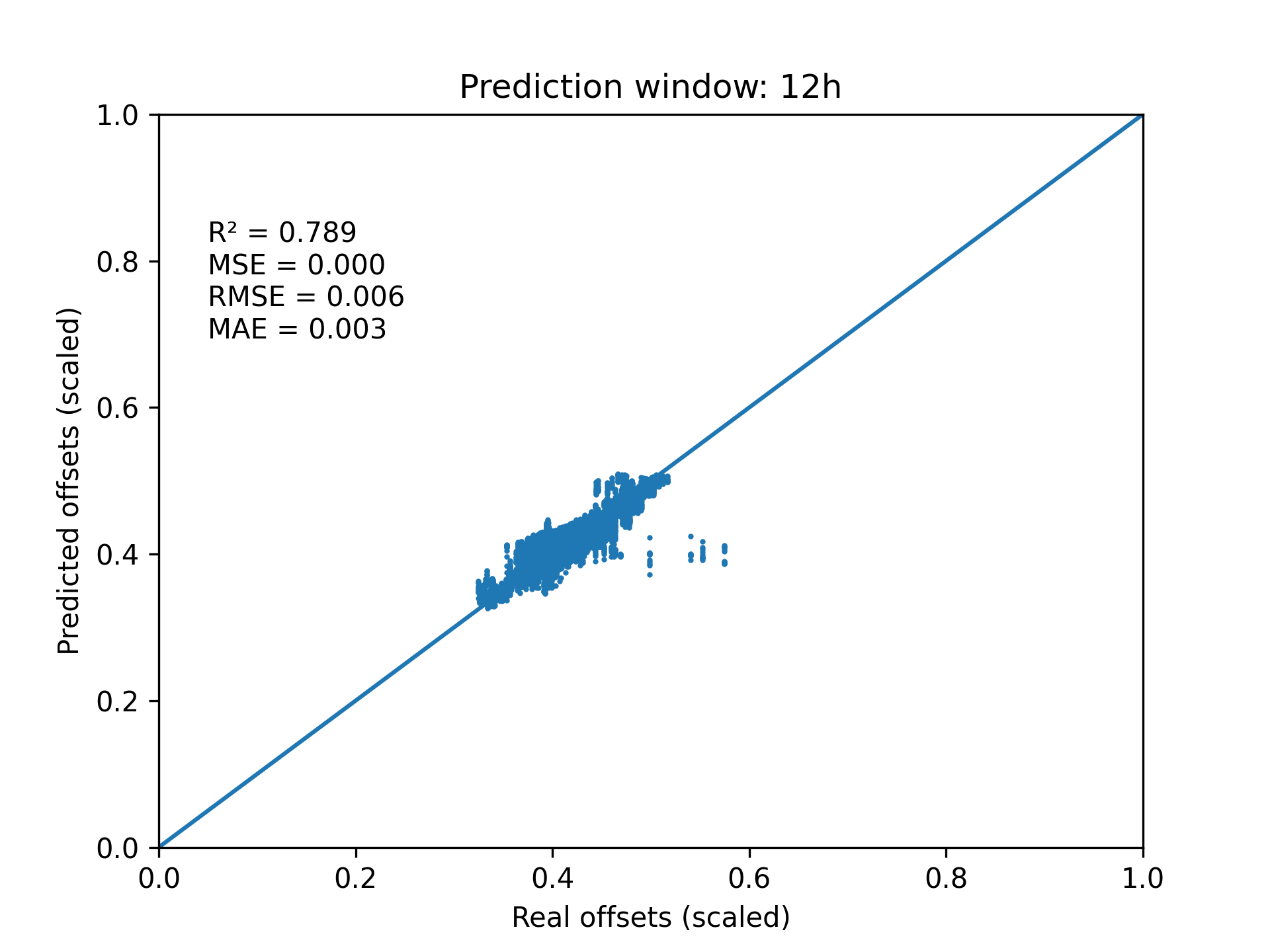}
         \caption{}
     \end{subfigure}
     \hfill
       \begin{subfigure}[b]{0.325\textwidth}
         \centering
         \includegraphics[scale=0.37]{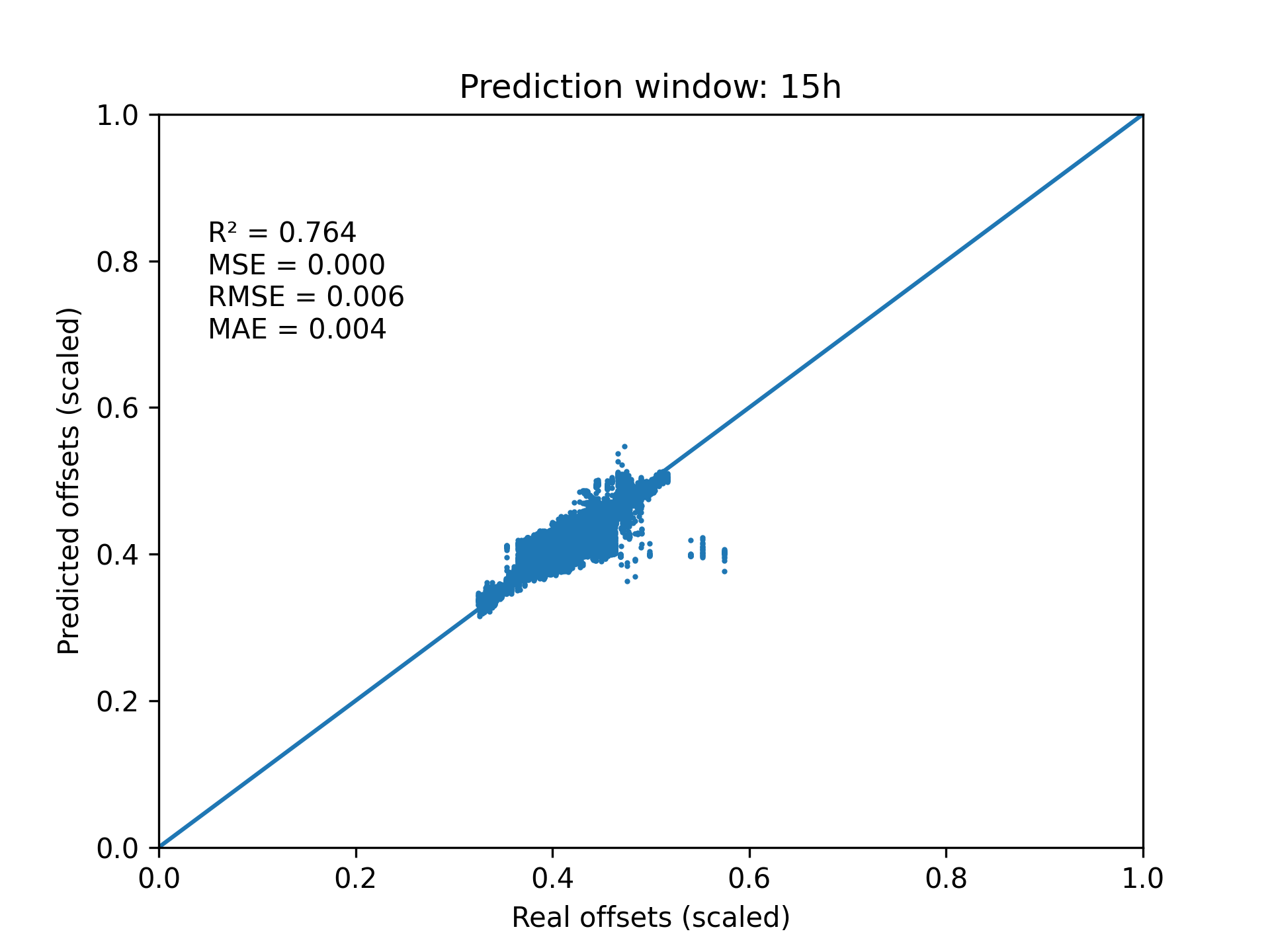}
         \caption{}
     \end{subfigure}
     \hfill
          \begin{subfigure}[b]{0.325\textwidth}
         \centering
         \includegraphics[scale=0.37]{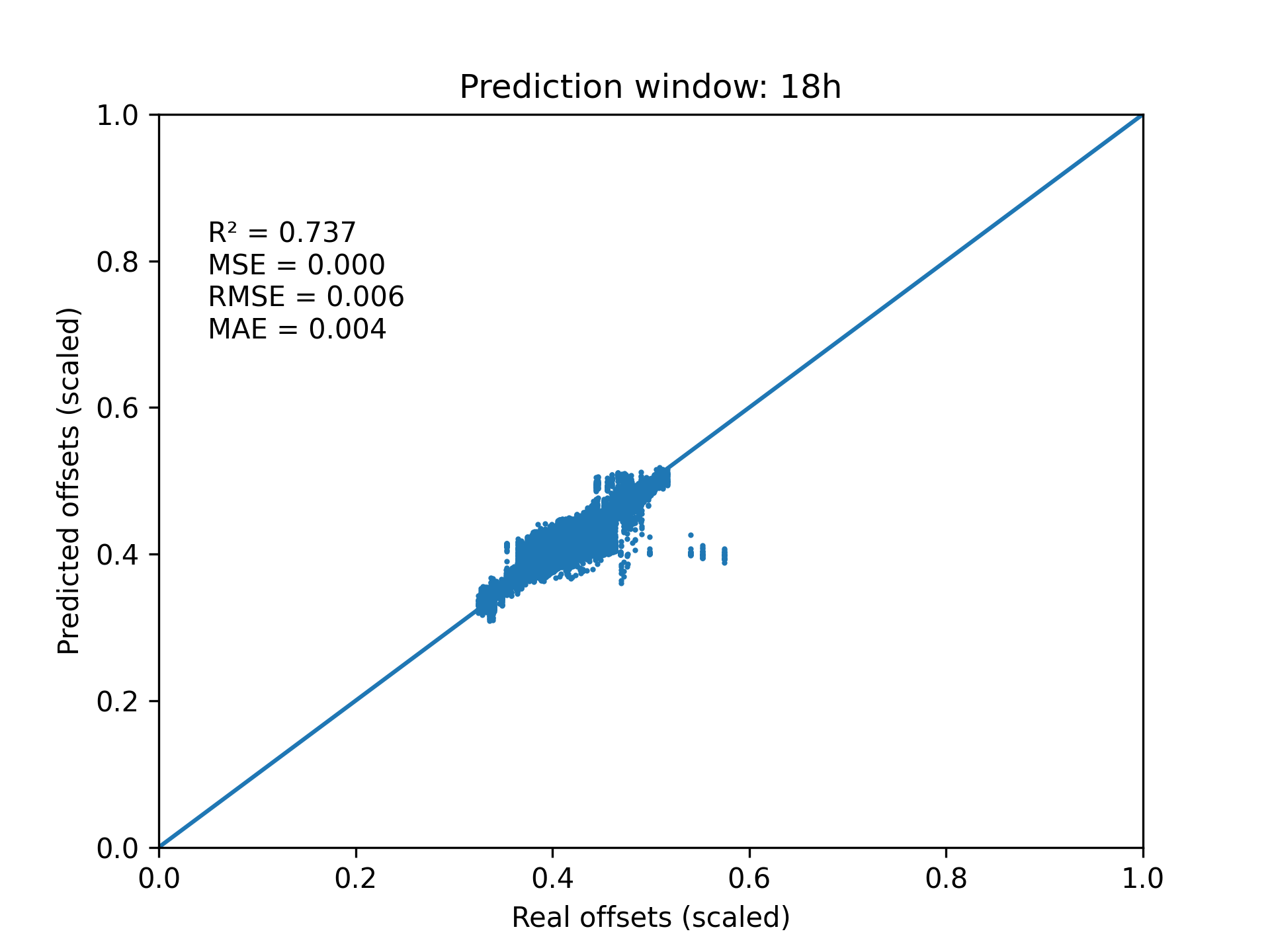}
         \caption{}
     \end{subfigure}
     \caption{Distribution of ML-predicted versus real offsets for different values of prediction window length, along with the $x=y$ line as a guide for the eye. An ideal prediction would correspond to all the data points lying on the $x = y$ line. The model that yielded the best performance was used in each case. Data are scaled as described in Section \ref{sec:Windowing}. The regression evaluation metrics for each case are also included in the respective subplot.}
     \label{fig:results_all_real_vs_pred_offsets}
\end{figure}

To evaluate the performance of the model in this scenario, the ML-predicted values of offsets, plotted as a function of their ``real" counterparts (as obtained from eq. \ref{eq:offsets}), for hurricane Ian and for different values of prediction window length, are presented in Figure \ref{fig:results_all_real_vs_pred_offsets}, with the respective evaluation metrics values included in Table \ref{tbl:results_all_real_vs_pred_offsets}. In these plots, the accuracy of the predictions is associated to the spread of the values around the $x=y$ line, with a perfect prediction corresponding to all values lying on the $x=y$ line. The evaluation metrics for the regression between the ML-predicted and the real offsets in each case are also shown in the respective plot. It is also worth mentioning that these results show the performance of the best model obtained for each prediction window. As in the previous scenario, the ML model is still capable of making highly accurate predictions ($R^2 > 0.9$). However, due to the use of a vastly larger training set, the ML model is still able to make reasonable predictions in larger prediction windows, with the best model yielding a value of $R^2 = 0.74$ for a prediction window of 18h.

\begin{table}[h!]
\caption{Evaluation metrics for the ML model in predicting offsets, for the different considered values of the prediction window.}
\label{tbl:results_all_real_vs_pred_offsets}
\begin{tabular}{ccccc}
 Prediction window (h) & $\text{R}^2$ & MSE & RMSE & MAE \\
\toprule
1 & 0.957 & <0.001 & 0.003 & 0.001 \\
3 & 0.922 & <0.001 & 0.003 & 0.002 \\
6 & 0.870 & <0.001 & 0.004 & 0.002 \\
9 & 0.827 & <0.001 & 0.005 & 0.003 \\
12 & 0.789 & <0.001 & 0.006 & 0.003 \\
15 & 0.764 & <0.001 & 0.006 & 0.004 \\
18 & 0.737 & <0.001 & 0.006 & 0.004 \\
\bottomrule
\\
\end{tabular}
\end{table}

\begin{figure}[h!]
     \begin{subfigure}[b]{0.01\textwidth}
         \centering
              \raisebox{1.3cm}{\rotatebox{90}{Prediction window: 1h}}
     \end{subfigure}
     \begin{subfigure}[b]{0.323\textwidth}
         \centering
         \includegraphics[scale=0.37]{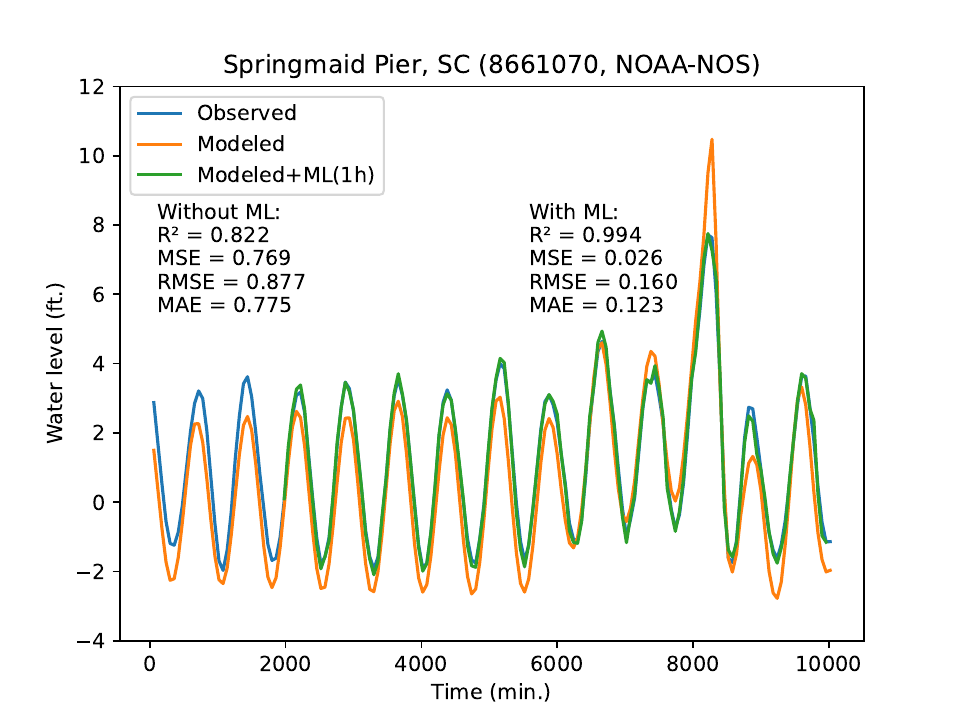}
         \caption{}
     \end{subfigure}
     \hfill
    \begin{subfigure}[b]{0.323\textwidth}
         \centering
         \includegraphics[scale=0.37]{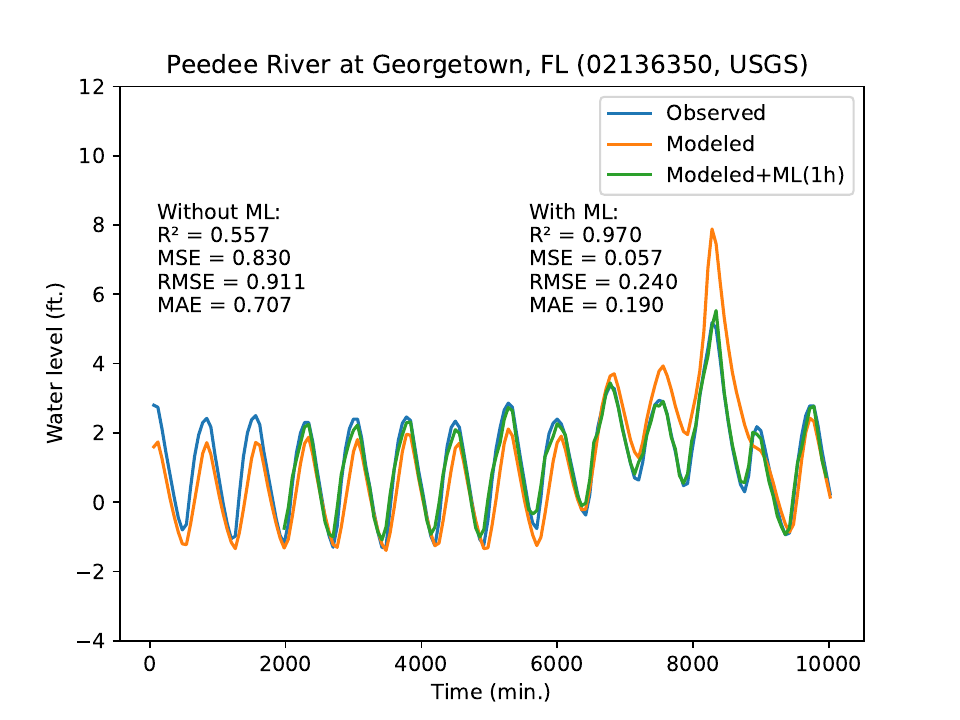}
         \caption{}
     \end{subfigure}
     \hfill
   \begin{subfigure}[b]{0.323\textwidth}
         \centering
         \includegraphics[scale=0.37]{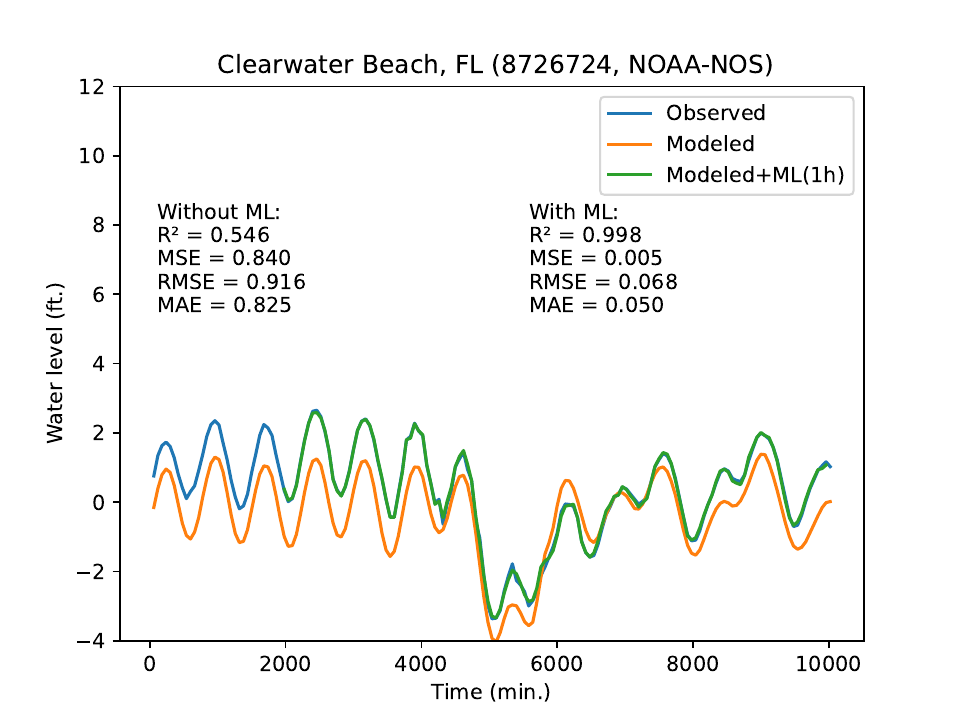}
         \caption{}
     \end{subfigure}
     \hfill
      \begin{subfigure}[b]{0.01\textwidth}
         \centering
              \raisebox{1.3cm}{\rotatebox{90}{Prediction window: 9h}}
     \end{subfigure}
        \begin{subfigure}[b]{0.323\textwidth}
         \centering
         \includegraphics[scale=0.37]{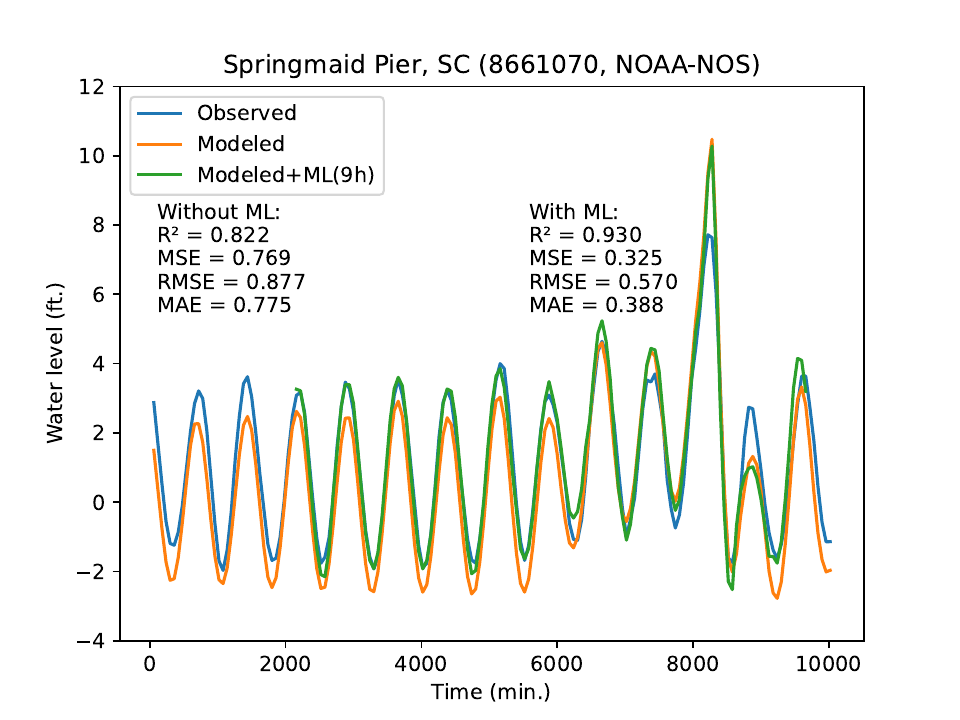}
         \caption{}
     \end{subfigure}
     \hfill
          \begin{subfigure}[b]{0.323\textwidth}
         \centering
         \includegraphics[scale=0.37]{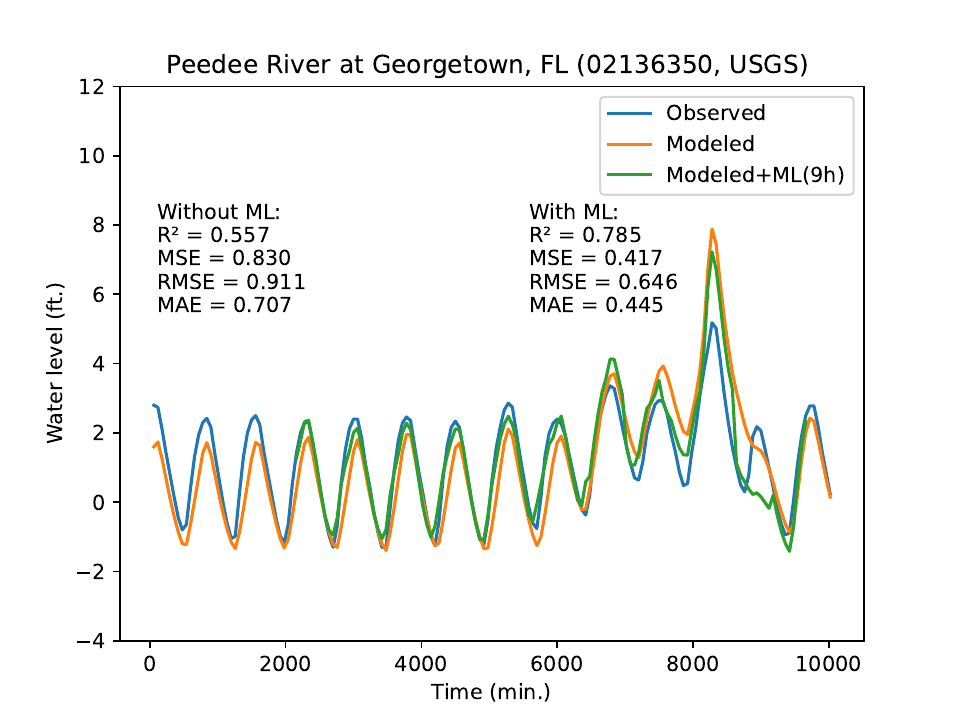}
         \caption{}
     \end{subfigure}
     \hfill
               \begin{subfigure}[b]{0.323\textwidth}
         \centering
         \includegraphics[scale=0.37]{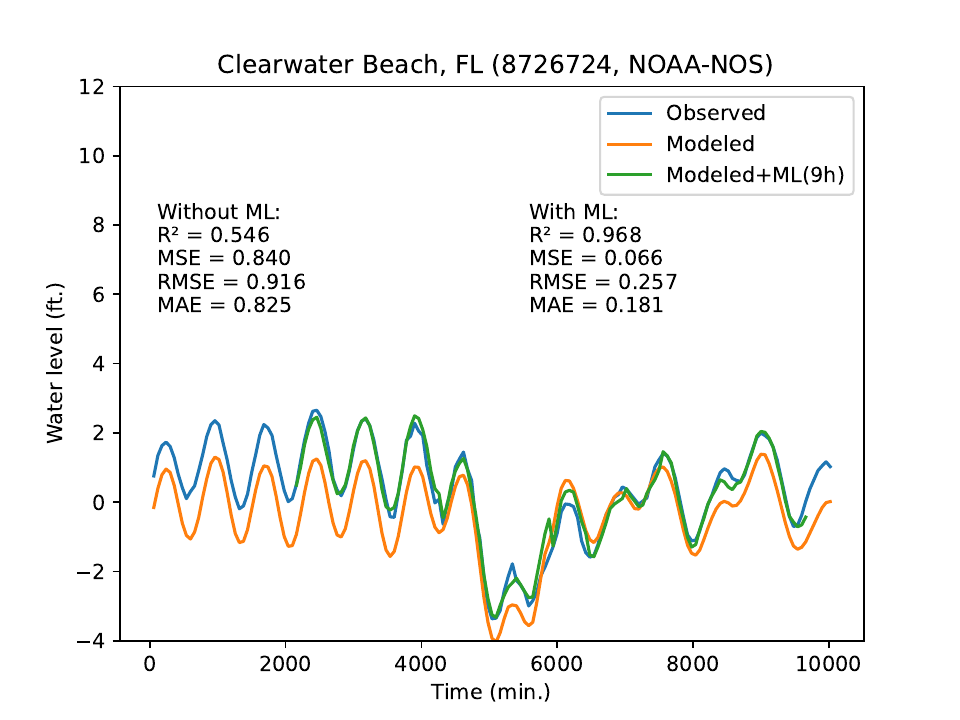}
         \caption{}
     \end{subfigure}
     \hfill
      \begin{subfigure}[b]{0.01\textwidth}
         \centering
              \raisebox{1.3cm}{\rotatebox{90}{Prediction window: 18h}}
     \end{subfigure}
      \begin{subfigure}[b]{0.323\textwidth}
         \centering
         \includegraphics[scale=0.37]{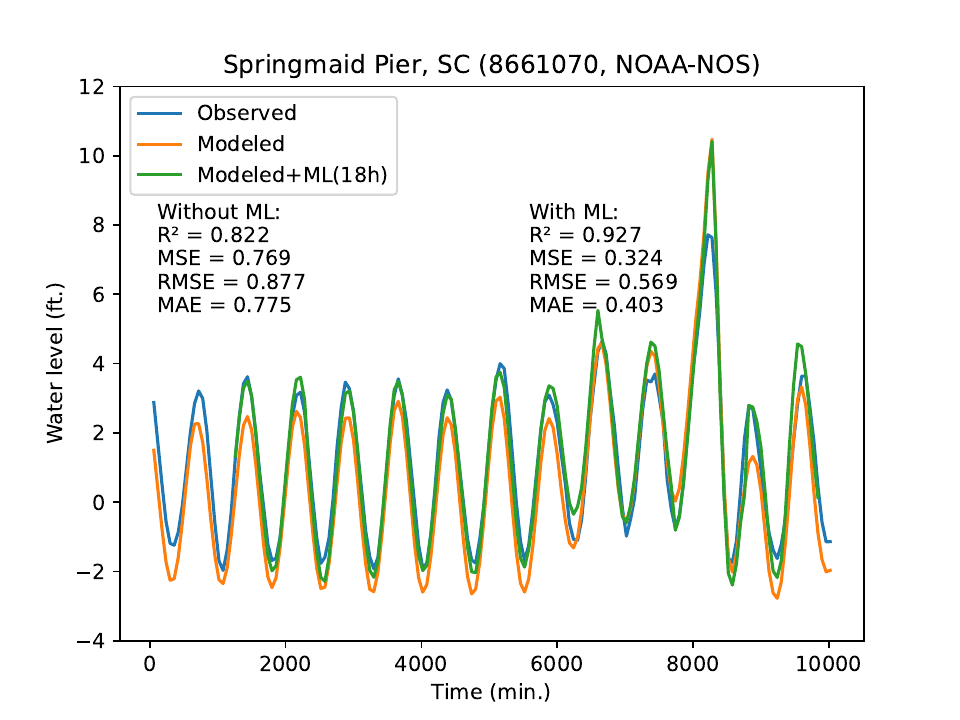}
         \caption{}
     \end{subfigure}
     \hfill
          \begin{subfigure}[b]{0.323\textwidth}
         \centering
         \includegraphics[scale=0.37]{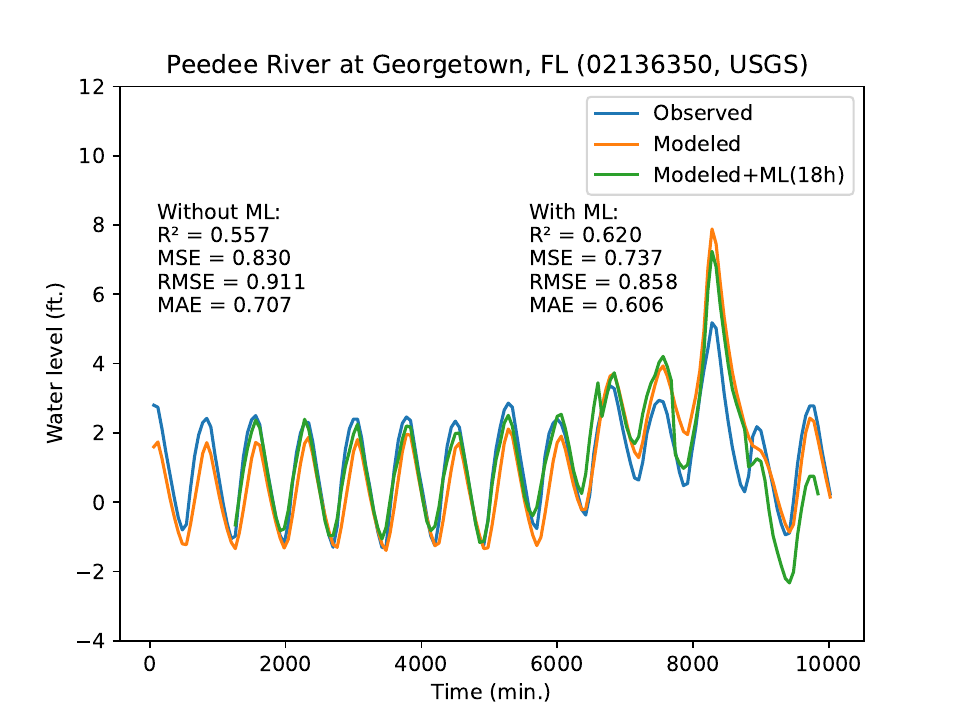}
         \caption{}
     \end{subfigure}
     \hfill
         \begin{subfigure}[b]{0.323\textwidth}
         \centering
         \includegraphics[scale=0.37]{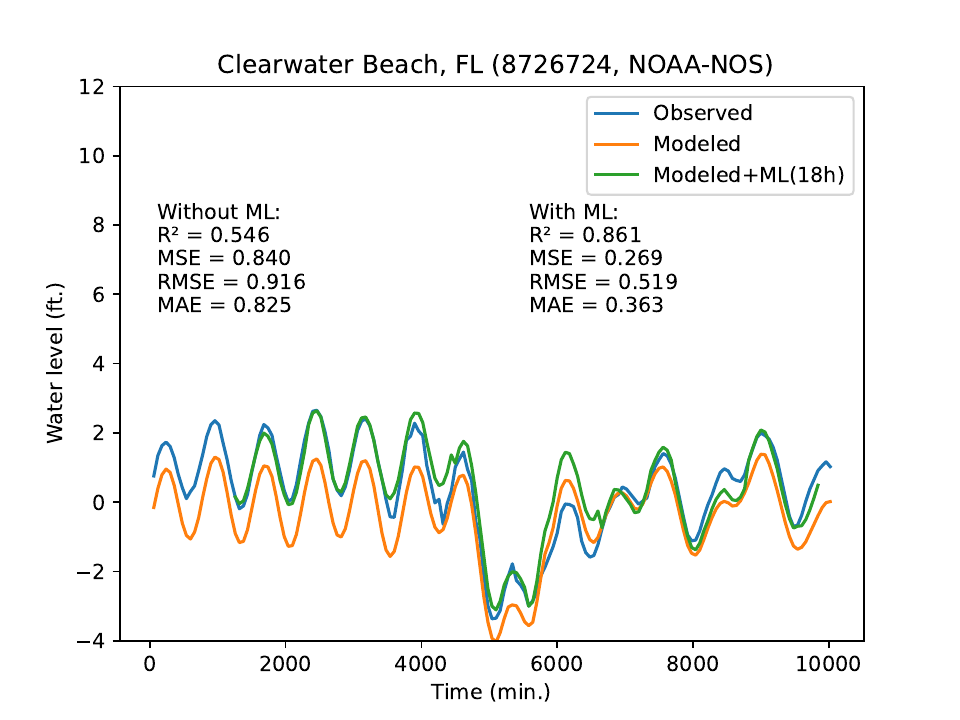}
         \caption{}
     \end{subfigure}
     \caption{Observed, (blue), modeled (orange) and ML-corrected modeled (green) values of water levels for 3 different gauge stations (columns) and 3 different sizes of the prediction window (1, 9 and 18 h, rows) from the test set (hurricane Ian, 2022). Evaluation statistics in each plot represent the performance of regression between modeled and observed water levels (left) and ML bias corrected and observed water levels (right). The ML model was trained on offsets data from all the available hurricanes in the considered dataset (\cite{CERA2023}). The model that yielded the best performance for each value of the prediction window was used in each case.}
     \label{fig:results_all_stations}
\end{figure}

To further examine the applicability of the proposed model for bias correction, we applied these corrections to the modeled water level values by subtracting the predicted offsets (shown in Figure \ref{fig:results_all_real_vs_pred_offsets}) from the ADCIRC-predicted water levels, and we compared these values (``Modeled+ML") with both the observed and the ADCIRC-predicted water levels without bias correction (''Modeled") in all gauge stations available in the hurricane Ian dataset. For an indicative demonstration, we selected three stations (shown in Figure \ref{fig:methods_cera_1}); Springmaid Pier and Peedee River at Georgetown, which are two, USGS and NOAA-NOS stations, respectively, in the vicinity of the second landfall of hurricane Ian, and Clearwater Beach, which is a NOAA-NOS station, relatively farther from the hurricane best track, but still at a relative proximity to the heavily inundated region of Lee County, FL, where hurricane Ian made its highly destructive first landfall. The observed, ADCIRC-modeled and ML bias corrected ADCIRC-modeled water levels in these three stations, for three different prediction window values, are shown in Figure \ref{fig:results_all_stations}. In the first two stations (shown in the first two columns in Figure \ref{fig:results_all_stations}), the landfall event is evident by the upward shift of the water levels time series, leading eventually to a pronounced ``spike" and then followed by a return to the normal tidal cycle. On the other hand, one of the main characteristics of the water level time series in the third station is the reduction of the water levels at the time of the landfall to the nearby region. The physics-based model used for the water level predictions tended to slightly underestimate water levels in most cases, but to overestimate them in the aforementioned cases with increased water levels due to hurricane landfall. Adding the ML-based bias correction (``Modeled+ML") evidently leads to an overall improvement of the prediction, as shown by the improved regression evaluation metrics in all cases (Figure \ref{fig:results_all_stations}). Despite the reduction of the predicting accuracy of the ML model at large prediction windows (e.g., 18h) in terms of offsets, as shown in Figure \ref{fig:results_all_real_vs_pred_offsets}, an improved prediction of water levels is still achieved when the results of the ML model are added back to the results of the physics-based model, with only the amount of this correction being reduced with increasing prediction window. Therefore, the observed limited predicting ability of the ML model at large prediction windows does not impact negatively the the combined "Modeled+ML" predicting accuracy, compared to the non-bias corrected "Modeled" results.

The predicted water levels with the ML bias correction model shown in Figure \ref{fig:results_all_stations} follow closely their observed counterparts in the majority of the various characteristic of each time series. Both normal tidal cycle and hurricane-related events, such as the aforementioned upward and downward ``spikes" in water levels, are being captured appropriately in most cases. The ML bias correction model appears limited only at increased prediction windows, and particularly in improving the prediction at the water level maxima in the first two stations; however in none of these cases the overall performance of the bias-corrected model was inferior than the non bias-corrected one.

\begin{table}[h!]
\caption{Evaluation statistics for the regression accuracy in water level prediction by the physics based model before and after applying the ML based bias correction model. Results are presented for the three selected stations and prediction window values used in Figure \ref{fig:results_all_stations}.}
\label{tbl:results_all_stations}
\begin{tabular}{lcccc}
\multicolumn{5}{c}{Springmaid Pier, SC (8661070, NOAA-NOS)}                                             \\ \hline
\begin{tabular}[c]{@{}l@{}}Prediction\\ window (h)\end{tabular} & \begin{tabular}[c]{@{}c@{}}R$^2$\\ (Without/With ML)\end{tabular} & \begin{tabular}[c]{@{}c@{}}MSE\\ (Without/With ML)\end{tabular} & \begin{tabular}[c]{@{}c@{}}RMSE\\ (Without/With ML)\end{tabular} & \begin{tabular}[c]{@{}c@{}}MAE\\ (Without/With ML)\end{tabular} \\ \hline
1 & 0.822 / 0.994 & 0.769 / 0.026 & 0.877 / 0.161 & 0.775 / 0.123 \\
9 & 0.822 / 0.930 & 0.769 / 0.325 & 0.877 / 0.570 & 0.775 / 0.388 \\
18 & 0.822 / 0.927 & 0.769 / 0.323 & 0.877 / 0.569 & 0.775 / 0.403 \\
\hline

    & \multicolumn{1}{l}{}   & \multicolumn{1}{l}{}  & \multicolumn{1}{l}{} & \multicolumn{1}{l}{} \\ 
\multicolumn{5}{c}{Peedee River at Georgetown, FL (02136350, USGS)}                                     \\ \hline
\begin{tabular}[c]{@{}l@{}}Prediction\\ window (h)\end{tabular} & \begin{tabular}[c]{@{}c@{}}R$^2$\\ (Without/With ML)\end{tabular} & \begin{tabular}[c]{@{}c@{}}MSE\\ (Without/With ML)\end{tabular} & \begin{tabular}[c]{@{}c@{}}RMSE\\ (Without/With ML)\end{tabular} & \begin{tabular}[c]{@{}c@{}}MAE\\ (Without/With ML)\end{tabular} \\ \hline
1 & 0.557 / 0.971 & 0.830 / 0.057 & 0.911 / 0.239 & 0.707 / 0.189  \\
9 & 0.557 / 0.785 & 0.830 / 0.417 & 0.911 / 0.646 & 0.707 / 0.445  \\
18 & 0.557 / 0.621 & 0.830 / 0.736 & 0.911 / 0.858 & 0.707 / 0.606  \\
\hline

 & \multicolumn{1}{l}{} & \multicolumn{1}{l}{}  & \multicolumn{1}{l}{}  & \multicolumn{1}{l}{} \\ 
\multicolumn{5}{c}{Clearwater Beach, FL (8726724, NOAA-NOS)}                                             \\ \hline
\begin{tabular}[c]{@{}l@{}}Prediction\\ window (h)\end{tabular} & \begin{tabular}[c]{@{}c@{}}R$^2$\\ (Without/With ML)\end{tabular} & \begin{tabular}[c]{@{}c@{}}MSE\\ (Without/With ML)\end{tabular} & \begin{tabular}[c]{@{}c@{}}RMSE\\ (Without/With ML)\end{tabular} & \begin{tabular}[c]{@{}c@{}}MAE\\ (Without/With ML)\end{tabular} \\ \hline
1 & 0.546 / 0.998 & 0.840 / 0.005 & 0.917 / 0.069 & 0.825 / 0.050  \\
9 & 0.546 / 0.968 & 0.840 / 0.066 & 0.917 / 0.258 & 0.825 / 0.181 \\
18 & 0.546 / 0.861 & 0.840 / 0.270 & 0.917 / 0.519 & 0.825 / 0.363 \\
\hline
\end{tabular}
\end{table}

\begin{figure}[h!]
     \centering
     \begin{subfigure}[b]{0.325\textwidth}
         \centering
         \includegraphics[scale=0.375]{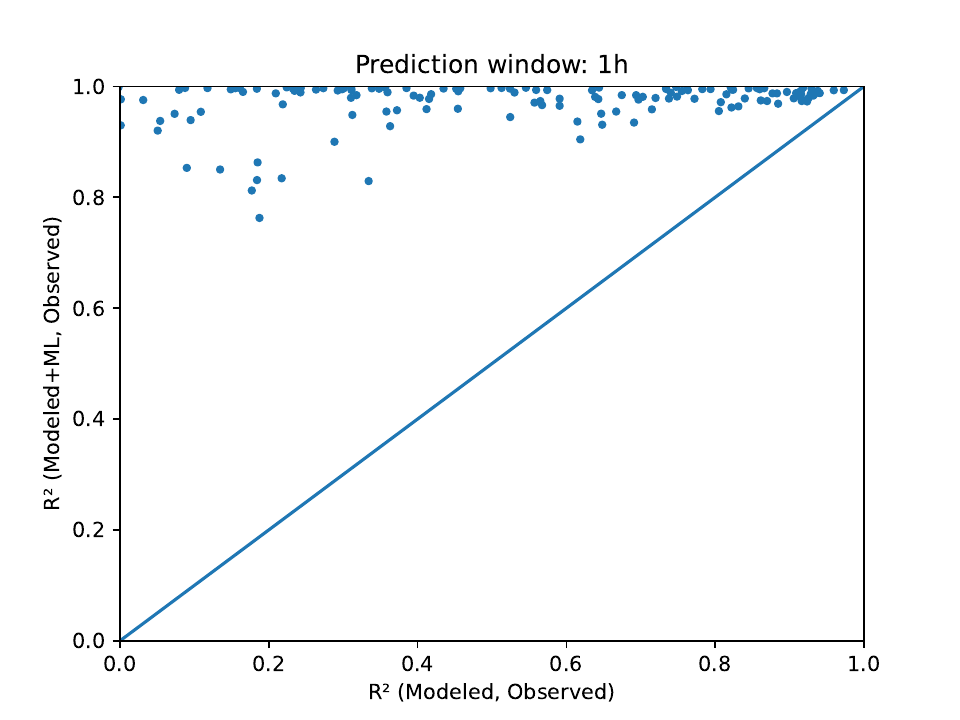}
         \caption{}
     \end{subfigure}
     \hfill
          \begin{subfigure}[b]{0.325\textwidth}
         \centering
         \includegraphics[scale=0.375]{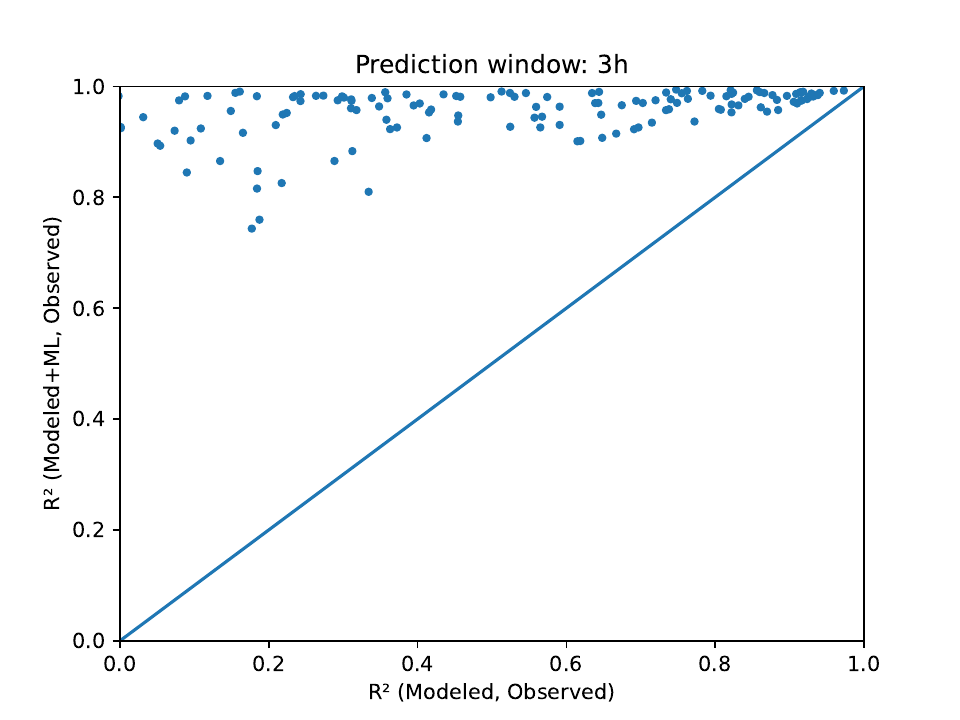}
         \caption{}
     \end{subfigure}
     \hfill
               \begin{subfigure}[b]{0.325\textwidth}
         \centering
         \includegraphics[scale=0.375]{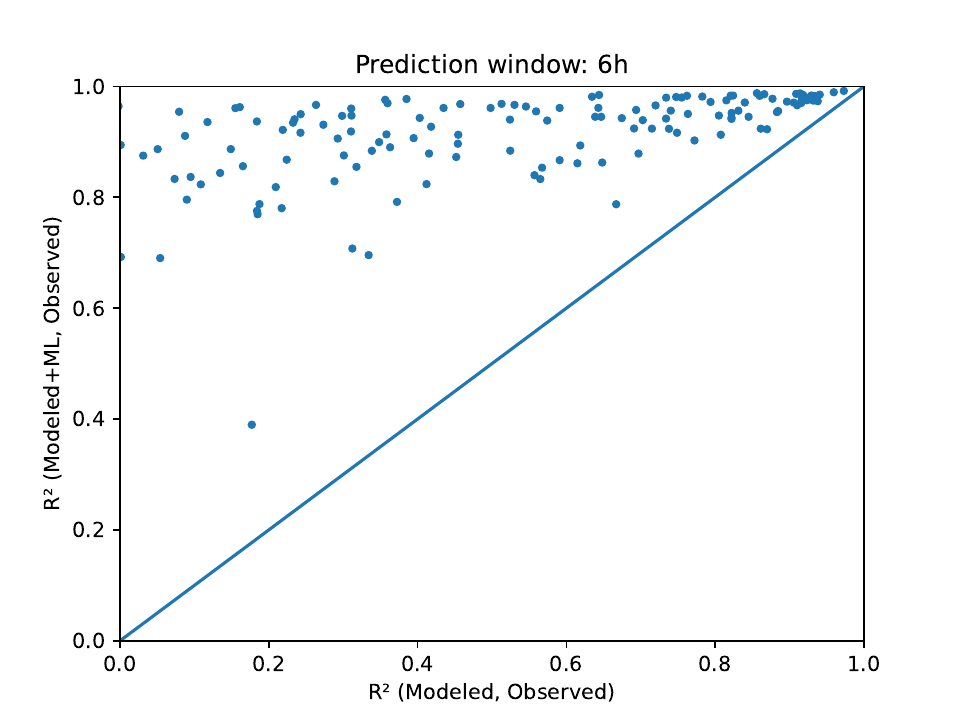}
         \caption{}
     \end{subfigure}
     \hfill
          \begin{subfigure}[b]{0.325\textwidth}
         \centering
         \includegraphics[scale=0.375]{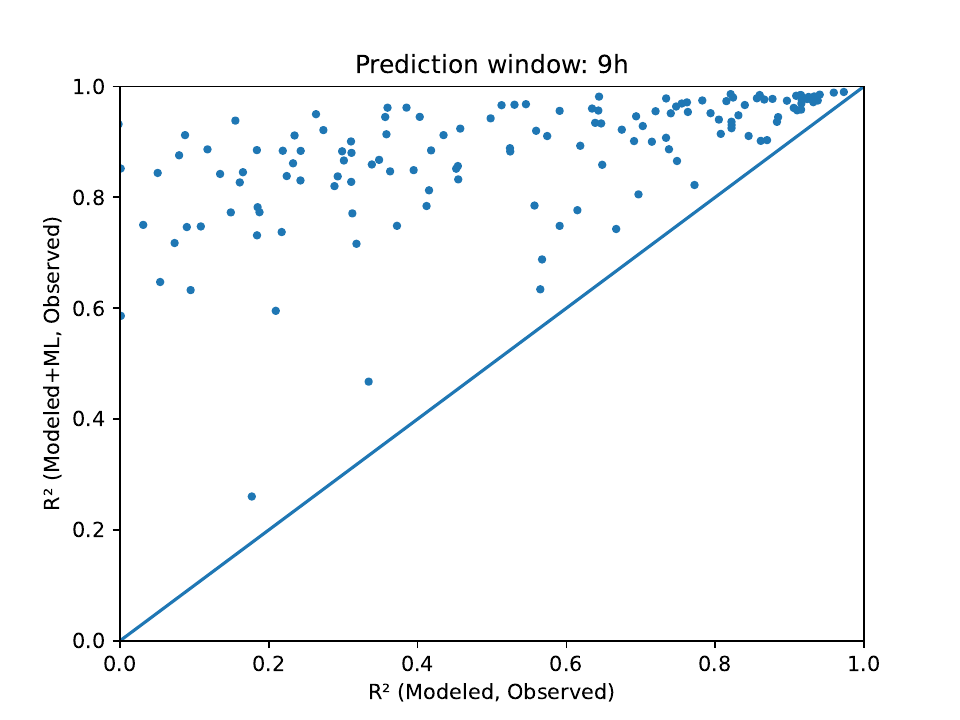}
         \caption{}
     \end{subfigure}
     \hfill
          \begin{subfigure}[b]{0.325\textwidth}
         \centering
         \includegraphics[scale=0.375]{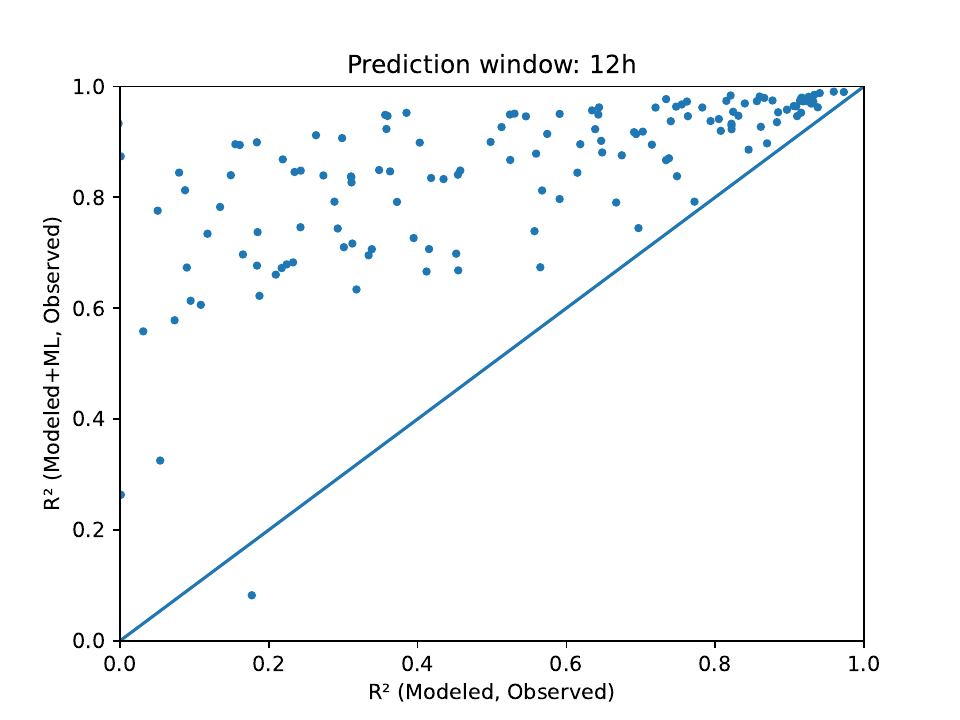}
         \caption{}
     \end{subfigure}
     \hfill
               \begin{subfigure}[b]{0.325\textwidth}
         \centering
         \includegraphics[scale=0.375]{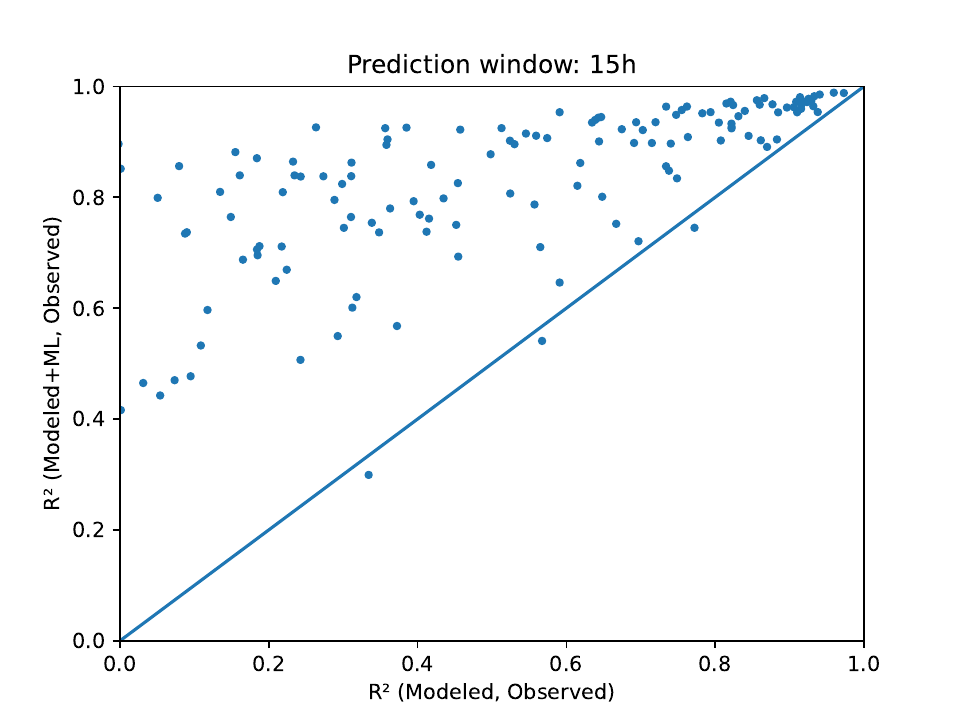}
         \caption{}
     \end{subfigure}
     \hfill
          \begin{subfigure}[b]{0.325\textwidth}
         \centering
         \includegraphics[scale=0.375]{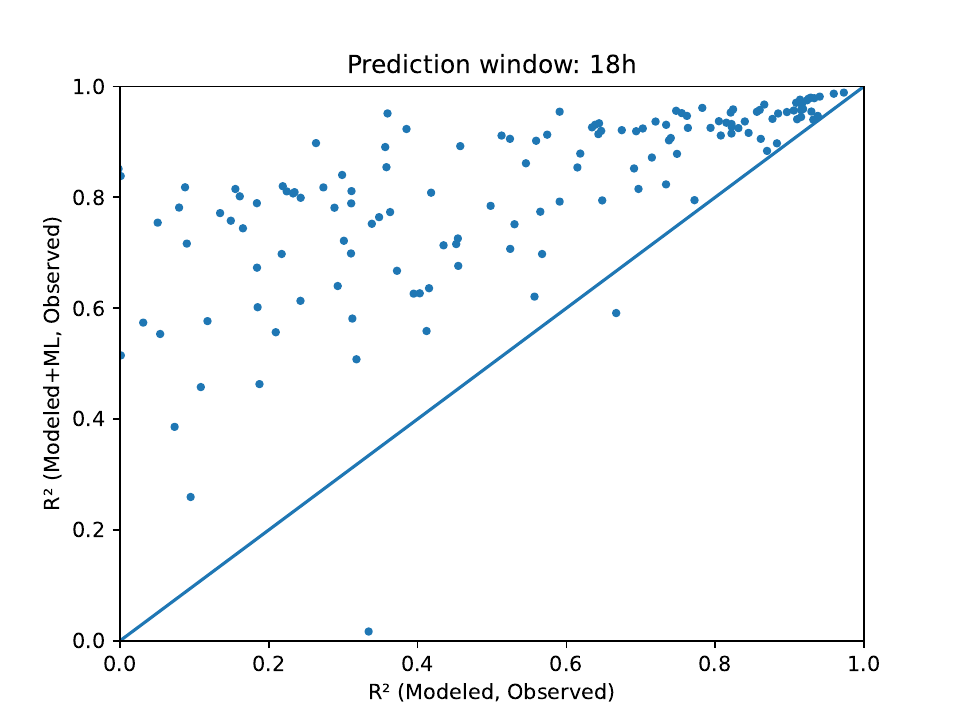}
         \caption{}
     \end{subfigure}
     \caption{Distribution of R$^2$ between ML bias corrected modeled water levels (``Modeled+ML") and their observed counterparts as a function of R$^2$ between modeled water levels before applying ML bias correction (``Modeled") and their observed counterparts, for all the available gauge stations in the hurricane Ian dataset (\cite{CERA2023}). Each plot corresponds to a different size of the prediction window (1-18h, a-g). The $x = y$ line is shown in each plot as a guide for the eye. Values on the upper left (lower right) triangle defined by the $x = y$ line and the plot margins correspond to gauge stations in which adding the ML bias correction to the results of the physics-based model led to improved (reduced) prediction accuracy. Results were produced with the ML model that yielded the best performance for each prediction window size.}
     \label{fig:results_all_real_vs_pred_r2}
\end{figure}

To generalize this conclusion, we further examined whether applying the ML bias correction to the results of the physics-based model had a positive or negative impact in all the considered gauge stations of our test set. To this end, the distribution of the R$^2$ values between bias corrected modeled and observed water levels as a function of the R$^2$ values between non-bias corrected modeled and observed water levels for all gauge stations in the test set and all the different considered prediction window lengths is presented in Figure \ref{fig:results_all_real_vs_pred_r2}. Each point on a plot corresponds to a gauge station of the test set, and by examining whether this points falls on the upper left or lower right triangle, as defined by the $x = y$ line and the plot margins, we can evaluate whether the added ML bias correction improves or deteriorates, respectively, the water levels predictions by the physics-based model (without bias correction). It is observed that the ML model is capable of improving predictions even on fringe cases, regardless the size of the prediction window. In most cases, even in stations in which the physics-based model was found to under-perform (i.e., points with low values on the x-axis), applying the proposed ML bias correction model leads to an improvement of the regression, reflected by a high value of the aforementioned points on the y-axis. Increasing the prediction window size leads inevitably to a reduction of this correction, as seen by the reduction of the y-coordinates of most data points with increasing prediction window size. However, even at the largest considered prediction window (18h), the vast majority of data points lie on the upper left triangle, signifying that the ML model leads to an improved prediction of water levels in the vast majority of the gauge stations on the test set. 

\subsection{Comparison between using different subsets of hurricanes for training}
In the previous section, we demonstrated the predicting ability of our proposed LSTM-based ML model for bias correction, with our model being trained on a vast dataset of 61 hurricanes from \cite{CERA2023} and tested on data from hurricane Ian, obtained from the same source. However, in this case, the training set involves data from hurricanes on a span of almost two decades, with highly inhomogeneous characteristics (intensity, areas affected etc.). Some other inhomogeneities of the dataset might also be related to factors such as lack of or less accurate data for older storms. Therefore, the first point we are exploring in this section is whether these inhomogeneities have a positive or negative impact to the predicting capability of the ML model. Another consideration about the size of the training dataset is related to the practical applicability of such an approach in real-time scenarios. Just by using of data exclusively from past hurricanes for the training set enables training a model once and then using the pre-trained model on real time. However, reducing training time by optimizing the interplay between training data size and model accuracy (if there is not a monotonously increasing relationship between the two) could potentially unlock more heuristically-oriented solutions, involving, for instance, training several ML models for different real-world scenarios. Hence, the second point of this analysis is to evaluate the relationship between dataset size, training time and predictive capability. To address these points, we trained our proposed LSTM-based ML model on 4 different subsets of the initial full 60-storms dataset from \cite{CERA2023} (see also Section \ref{ssec:data} and Table \ref{tbl:scenarios}, scenarios 3-6). In these scenarios, we evaluated the accuracy of the model when trained on offsets data from hurricanes with either relatively similar or different tracks, compared to hurricane Ian. Offsets data from the later were again used as test set for evaluation in all scenarios. 

\begin{figure}[h!]
     \centering
     \centering
     \includegraphics[scale=0.75]{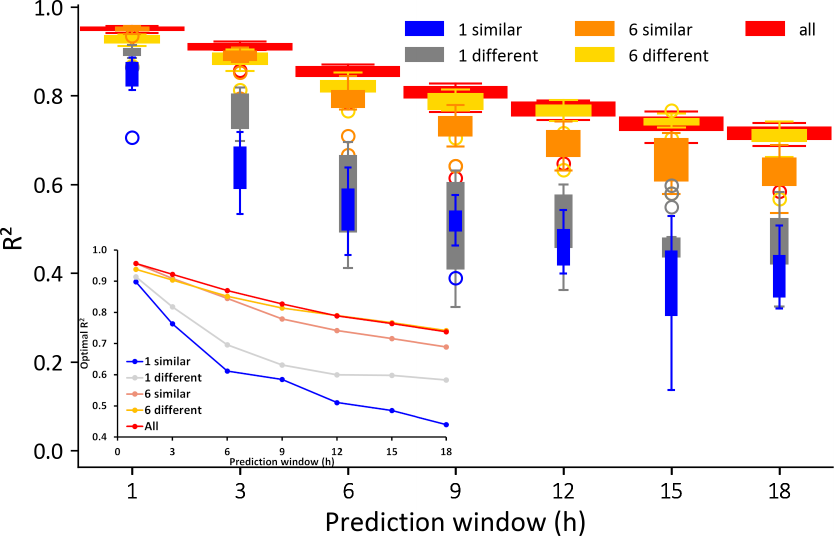}
     \caption{Distribution of R$^2$ values of the ML model in predicting offsets due to varying input window length, as a function of prediction window size, for the cases of using 1 similar (blue), 1 different (grey), 6 similar (orange) and 6 different (yellow) hurricanes compared to hurricane Ian for training, and hurricane Ian for testing the ML model. The width of each box is roughly proportional to the size of the training dataset for each case. The best R$^2$ values for each prediction window size are shown in the inset plot.}
     \label{fig:results_comparison_1}
\end{figure}

The distribution of the different values of the R$^2$ metric, obtained by using different values of the input window length in scenarios 2-6 (see Table \ref{tbl:scenarios}), as a function of prediction window length, is presented as a boxplot in Figure \ref{fig:results_comparison_1}. The highest (optimal) R$^2$ obtained in each case is shown in the inset figure. It is observed that the best performance in predicting offsets is achieved by using either the ``all" or the ``6 different" training datasets. From the inset of Figure \ref{fig:results_comparison_1}, which shows the highest R$^2$ value achieved in each scenario by varying input window length, using ``all" data is shown to lead to a slightly better performance in this particular case. However, the distributions of R$^2$ in these two scenarios appear highly overlapping in the corresponding boxplot in Figure \ref{fig:results_comparison_1}. Thus, we performed a Wilcoxon test (\cite{Wilcoxon1945}) to examine whether these distributions can be assumed to be drawn from the same distribution with statistical significance. The Wilcoxon test was chosen as the distribution of the considered data was not found to follow the normal distribution (\cite{Wilcoxon1945}). The results in Table \ref{tbl:Results-comparison-wilcoxon} verify the observation that the R$^2$ distributions in the "6-different" and the "all" scenarios are highly overlapping, as the null hypothesis that the two distributions are not significantly different is accepted for a prediction window greater of 12h and greater. Hence, it can be deducted that the performance of the proposed ML model is statistically similar for large prediction windows, regardless if it was trained on all available data or only on an almost random selection of 6 hurricanes, which is roughly $1/7$ of the size of the full dataset. Regarding the other scenarios, it is observed that the relevance of the training dataset to the target storm does not impact noticeably the predicting accuracy of the ML model. The slightly improved performance obtained with the ``6 different" dataset compared to the ``6 similar" one can be attributed to the greater size of the former compared to the latter (which is $\sim 1.67$ times larger). 

One of the parameters that impact the performance of the model, as shown in previous sections, is the size of the input window (Figure \ref{fig:sliding_window}). The overall trend can be seen from the analysis of scenario 1 (Figure \ref{fig:R2_Ian_1}), and the distribution of R$^2$ due to the variance induced by varying input window size in Figure \ref{fig:R2_Ian_2}. Using the different training datasets involved in scenarios 2-6 leads to changes in this relationship. While the overall trend is similar to the one in Figure \ref{fig:R2_Ian_1}) for all scenarios, the distribution of R$^2$ due to varying input window is different on each scenario, as shown in Figure \ref{fig:results_comparison_1}. In scenarios involving less data (``1 similar", ``1 different"), it is evident that the variance of R$^2$ due to varying input window length is greater, and tends to be increased with increasing prediction window length. However, on scenario 2 (``all"), but interestingly, also on scenario 6 (which is the one with the second greatest number of data points), this variance is noticeably smaller and invariant to the length of the prediction window. Therefore, using more data reduces the impact of optimizing the length of the input window.

\begin{figure}[h!]
     \centering
     \centering
     \includegraphics[scale=0.75]{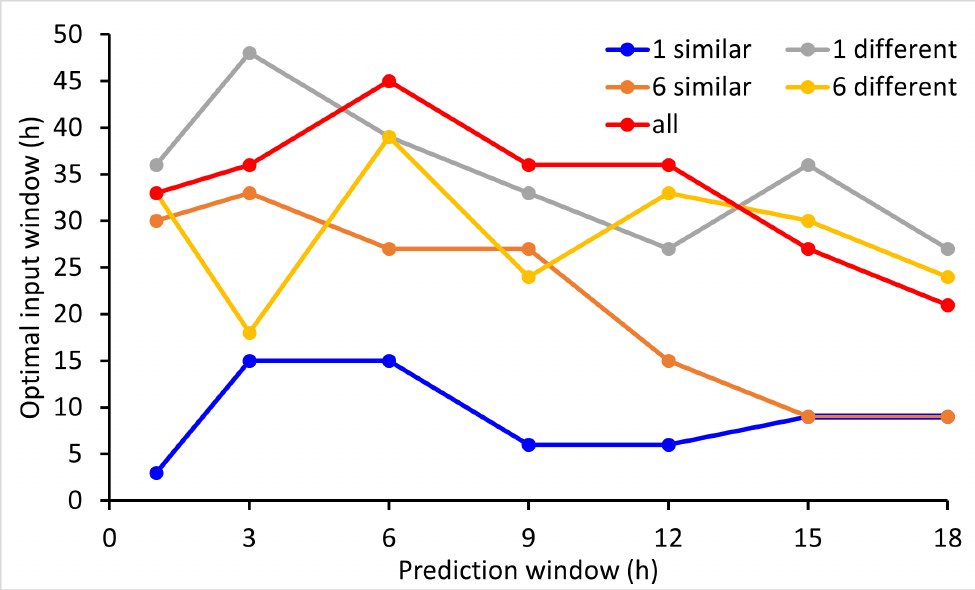}
     \caption{Length of the input window required for achieving the optimal performance of the model as a function of prediction window size for the same cases and color code as in Figure \ref{fig:results_comparison_1}.}
     \label{fig:results_comparison_2}
\end{figure}

The optimal size of the input window for achieving the best R$^2$ in predicting offsets in the hurricane Ian test set is shown in Figure \ref{fig:results_comparison_2}. Using hurricanes with relatively similar tracks to the target hurricane (``1 similar", ``6 similar"), leads to reduced input window sizes for optimal performance at large prediction windows. However, the seer performance of the model appears to be mostly dependent on the size of the training dataset, rather than its relevance to the target storm, as discussed above (Figure \ref{fig:results_comparison_1}). Moreover, as the use of more data was found to reduce the impact of optimizing the input window, using smaller input windows could potentially lead to a non significant performance loss in the cases of the ``6 different" and ``all" training sets.

\begin{table}[h!]
\caption{Wilcoxon rank test statistic and the corresponding p-values for the null hypothesis that the distributions of R$^2$ values obtained with the ``all" and ``6 different" training datasets (scenarios 2 and 6, respectively) come from the same distribution.}\label{tbl:Results-comparison-wilcoxon}
\begin{tabular}{ccc}
Prediction window (h) & Wilcoxon statistic & p-value\\
\toprule
1  & 0  & <0.001 \\
3  & 0  & <0.001 \\
6  & 1  & <0.001 \\
9  & 3  &  0.001 \\
12 & 40 &  0.159 \\
15 & 53 &  0.464 \\
18 & 39 &  0.144 \\
\bottomrule
\\
\end{tabular}
\end{table}

\begin{figure}[h!]
     \centering
     \begin{subfigure}[b]{0.475\textwidth}
         \centering
         \includegraphics[scale=0.975]{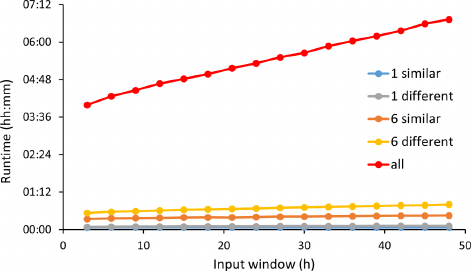}
         \caption{}
         \label{fig:results_comparison_runtime_1}
     \end{subfigure}
     \begin{subfigure}[b]{0.475\textwidth}
         \centering
         \includegraphics[scale=0.975]{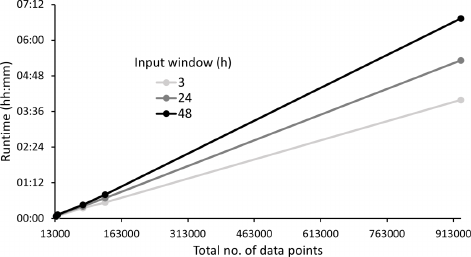}
         \caption{}
         \label{fig:results_comparison_runtime_2}
     \end{subfigure}
     \caption{(a) Computational time required to train the proposed ML model as a function of input window for the five different considered training datasets. (b) Runtime as a function of dataset size, plotted for three different sizes of the input window. All models were trained using 4 NVidia® A-100® GPUs.}
     \label{fig:results_comparison_runtime}
\end{figure}

The impact of the size of the training dataset and input window size to the computational time required for training the ML model in each case is demonstrated in Figure \ref{fig:results_comparison_runtime}. The runtime required for training the ML model when using subsets of the full ``all" training dataset can be reduced by roughly 3 to 7 times (Figure \ref{fig:results_comparison_runtime_1}). Moreover, the benefit of using small input windows is also evident, as it leads to a linear increase in runtime, which is most pronounced on the model trained on the full dataset (Figure \ref{fig:results_comparison_1}). The relationship between the training time and the number of data points in the training set appears to be linear, as expected, with the input window affecting the slope of this curve, as increasing size of the input window leads to an increase of the slope in the corresponding runtime vs dataset size curve (Figure \ref{fig:results_comparison_runtime_2}). In combination with the previous findings, it is thus evident that using the ``6 different" training subset can lead to offset predictions without significantly less accuracy than by using all available data (at least for large prediction windows), but at a 3 to 7 times reduced computational time required for the training of the ML model. 

\section{Conclusions}\label{sec:conclusions}
In this work, we propose and analyze the use of an LSTM-based ML architecture for capturing and predicting the behavior of the systemic error for storm surge forecast models with respect to real-world water height observations from gauge stations during hurricane events. The overall goal of this work is to predict the systemic error of the physics model and use it to correct the simulation results \textit{post factum} (i.e., to correct the model bias). The dataset of \cite{CERA2023} was used for the training and evaluation of the proposed ML model. In this dataset, ADCIRC (\cite{luettich1992, Dietrich2011, Westerink1992}) was used as the physics-based storm surge forecast model. Observed water level data were obtained from both \cite{NOAA} and \cite{USGS} gauge stations. This work demonstrated the advantages of a deep learning, LSTM-based framework for bias correction.

Offset times series data were systematically extracted from all available hurricanes in the dataset, calculated as the difference between observed and modeled water levels (see Eq. \ref{eq:offsets}), and subsequently, they were pre-processed as model inputs by the sliding windows approach. All data were also standardised and divided into training and validation datasets before model training. Hurricane Ian (2022) was chosen as the focal point of this work, due to its severity and massive impact in the Southeast U.S., thus it was considered as the test case, with its data being kept fully unknown to the ML model during its training.

The proposed model showed a promising capability in predicting the offset values, with $\text{R}^2 > 0.9$ for small prediction windows and $\text{R}^2 \sim 0.7 - 0.8$ for increased prediction windows up to 18h into the future (see Figures \ref{fig:results_all_real_vs_pred_offsets} and \ref{fig:results_comparison_1}). In all cases, adding the ML predicted offsets back to the modeled water levels ("Modeled+ML") was found to lead to an improvement in predicting the observed water levels for the vast majority of the gauge stations in the test dataset, in comparison with the performance of the physics based model prior to the application of the ML bias corrections ("Modeled", see Figures \ref{fig:results_all_stations} and \ref{fig:results_all_real_vs_pred_r2}).

Moreover, we analyzed the performance of the ML bias correction model trained with different subsets of the full 61 hurricanes initial dataset, and we observed that by using a training subset of 6 hurricanes with different tracks compared to hurricane Ian, with a size of roughly 1/7th of the full dataset, the performance the model was not statistically different to the performance of the model trained with the full dataset, for prediction windows of 9h and greater. The similarity in the performance of the model in these two cases was also verified by the results of a Wilcoxon test, with the performance values of the two models in terms of $\text{R}^2$ for different values of input windows as input distributions.

Using similar storms to the target storm can reduce training times, as shown by the reduced input window required for achieving optimal performance in the cases where similar storms (to the target one) were used (see Figure \ref{fig:methods_cera_2}), and the computational time required for training was found to scale linearly with increasing input window (Figure \ref{fig:results_comparison_runtime_1}). However, in real scenarios, using similar storms requires the a priori knowledge of the storm track, which limits applicability in operational situations. Training time was also found to increase linearly with the number of data points in the training set, as expected (see Figure \ref{fig:results_comparison_runtime_2}).

There are a few limitations of the applied methodology uncovered in our study that can motivate further investigation. Operational forecasting guidelines can require forecasting windows up to 3 days (72h). Since this magnitude is almost half of the duration of the data for each hurricane in the dataset used in this work, assessing the performance of our ML bias correction model for such periods in this context is not yet possible. Future work involves the extension of this dataset and also the use of different ML architectures, with bidirectional and/or self-attention based LSTM, combined with CNN as potential candidates. The ML-based geospatial extrapolation of bias correction values to the entire simulation area beyond the gauge stations poses another interesting topic.

In summary, we demonstrated and analyzed the use of an LSTM-based framework for predicting systemic errors in storm surge physics-based forecast models. Our approach showcased promising results in predicting offsets, providing a pathway for enhancing forecast models as a post processing bias correction component. By selecting to use only offsets data from past hurricanes for training, we ensure that this approach is highly applicable in operational situations and also highly transferable to other forecasting models. 

\section{Acknowledgments}
We acknowledge the support of the Department of Energy (DoE) through the award DE-SC0022320 (MuSiKAL). We would also like to thank Louisiana State University (LSU) and the Center for Computation and Technology at LSU for granting allocations for their computing resources and storage space.

\bibliographystyle{unsrtnat}
\bibliography{ms}

\end{document}